\documentclass[11pt]{article}
\usepackage{amsthm}
\usepackage[final]{acl}

\usepackage{newtxtext,newtxmath}
\usepackage{microtype}
\usepackage{graphicx}
\usepackage{makecell}
\usepackage{stfloats}
\usepackage{booktabs} 
\usepackage{float} 
\usepackage{alltt}   
\usepackage{minitoc} 
\usepackage{etoc} 
\usepackage{hyperref}

\usepackage{mathtools}
\usepackage{multirow}
\usepackage{blindtext}
\usepackage{bm}
\usepackage[dvipsnames]{xcolor}

\usepackage{caption}
\usepackage{subcaption} 
\usepackage[toc,page,header]{appendix}
\usepackage{minitoc}

\title{From Detection to Refusal: Safer LLMs via Circuit-Guided Weight Scaling}

\author{
  \textbf{Kuan-Lin Chu}$^{*\dagger}$,
  \textbf{Chung-En Sun}$^{*\dagger}$,
  \textbf{Tsui-Wei Weng}$^\ddagger$ \\
  $^\dagger$Computer Science and Engineering, University of California San Diego \\
  $^\ddagger$Halıcıoğlu Data Science Institute, University of California San Diego \\
  \textnormal{\{klinchu, cesun, lweng\}@ucsd.edu}
}
\begin{document}

\maketitle
\begingroup
\renewcommand\thefootnote{}\footnotetext{$^*$Equal contribution.}
\endgroup
\begin{abstract}
Despite extensive alignment efforts, Large Language Models (LLMs) remain vulnerable to generating unsafe content under adversarial prompting, yet the internal mechanisms by which safety behaviors are implemented remain poorly understood. We study LLM safety from a mechanistic interpretability perspective and characterize a multi-stage \emph{safety circuit} that organizes refusal behavior, consisting of (i) \textbf{Harmful Detection Heads} that respond to harmful inputs, (ii) \textbf{Safety Neurons} that mediate and stabilize safety signals in the residual stream, and (iii) \textbf{Refusal Heads} that translate these signals into safe response generation. Using targeted attention-head and neuron-level interventions, we provide causal evidence consistent with this circuit organization, showing that suppressing upstream Harmful Detection Heads disrupts downstream refusal behavior and that safety neurons mediate this interaction. We validate that this decomposition recurs across multiple LLM architectures and adversarial attack settings, and use simple, architecture-preserving weight scaling as a mechanistic probe to test its functional relevance. Across six LLMs, circuit-guided scaling improves safety rates under attacks by 26.5\%, while incurring only a 1.7\% accuracy drop across four standard benchmarks. Overall, our results support a circuit-level interpretation of LLM safety and suggest that mechanistic abstractions can reveal stable and transferable patterns underlying aligned behavior. Code is available at \url{https://github.com/Trustworthy-ML-Lab/Detection2Refusal}.
\end{abstract}

\doparttoc 
\faketableofcontents 

\section{Introduction}
\label{sec:introduction}

Large language models (LLMs) have achieved remarkable performance across a wide range of applications, including conversational assistants, code generation, and content creation. Despite these advances, LLMs remain vulnerable to producing harmful or unsafe outputs, particularly under adversarial prompting~\citep{harmful}. While substantial effort has been devoted to mitigating these behaviors, a fundamental question remains largely unanswered: \emph{how are safety behaviors internally implemented within LLMs?} Addressing this question is essential not only for interpreting model behavior, but also for understanding the structural limits and failure modes of existing safety approaches.

Most existing safety methods operate at the behavioral or semantic level. Techniques such as reinforcement learning from human feedback (RLHF)~\citep{RLHF}, adversarial training~\citep{Adv-training}, and post-hoc filtering or moderation~\citep{prompt-filtering} aim to shape model outputs without explicitly characterizing the internal mechanisms that detect harmful intent or trigger refusal. Concept-based approaches, such as Concept Bottleneck LLMs~\citep{cbllm}, introduce interpretable intermediate representations, but still rely on externally defined abstractions rather than uncovering how safety is realized within the model’s native computation. As a result, these methods provide limited insight into \emph{where} and \emph{how} safety decisions are made inside the network.

Recent progress in mechanistic interpretability has shown that high-level behaviors in LLMs can often be localized to specific internal components, such as attention heads or individual neurons~\cite{inductionheads,safe_heads,safety_neurons}. In the context of safety, prior work has identified components correlated with harmful content detection or refusal behavior, and has shown that editing a small set of such heads can strengthen refusal~\citep{DRefA, safe_heads}. However, these studies primarily focus on component attribution in isolation, and do not characterize how multiple components interact to jointly support safety behavior. As a result, it remains unclear whether safety arises from coordinated internal dependencies or from independent mechanisms.

In this work, we take a mechanistic interpretability perspective on LLM safety and characterize a \emph{detection--refusal} circuit-level organization of refusal behavior that recurs across models. Rather than treating safety-related heads and neurons as isolated features, we show that safety behavior can be usefully decomposed into three interacting, layer-stratified roles: 
(i) \textbf{Harmful Detection Heads} that respond selectively to harmful inputs, localized primarily across the early and middle layers of the network;
(ii) \textbf{Safety Neurons} whose activations modulate the strength of refusal-related signals; and 
(iii) \textbf{Refusal Heads} that translate these signals into safe or refusing tokens. 
Crucially, these components operate across a macro-level layer hierarchy: the early-stage detection heads compute upstream features that causally drive the activation of the late-stage safety neurons and refusal heads. Across multiple architectures, we provide causal intervention evidence consistent with this cross-layer organization, showing that removing the early detection heads or mid-to-late safety neurons directly weakens downstream refusal-head activity.

To leverage these mechanistic insights in practice, we apply a simple, architecture-preserving weight-scaling intervention that reinforces the identified components without additional training or optimization. Scaling each component's impact through weight parameters individually (detection heads, safety neurons, or refusal heads) consistently improves safety, while jointly scaling them yields substantially larger gains than any single intervention alone. Averaged across six LLMs, this training-free circuit-guided scaling improves safety rates under attacks by 26.5\% (from 43.2\% to 69.7\%), while largely preserving the model’s original utility, with an average accuracy reduction of only 1.7\% across four task-oriented benchmarks.

Our contributions are summarized as follows:
\begin{itemize}
    \item We characterize a \textbf{detection--refusal circuit-level organization} of refusal behavior in LLMs---consisting of \emph{Harmful Detection Heads}, \emph{Safety Neurons}, and \emph{Refusal Heads}---and validate this interpretable decomposition with \textbf{causal intervention evidence}, demonstrating that selectively zeroing out detection heads directly weakens downstream refusal-head activity.
    \item Leveraging the mechanistic insights from the detection-refusal circuits, we show that \textbf{simple circuit-guided weight scaling} substantially improves safety: scaling each component weight individually increases robustness, while jointly scaling them yields larger gains, improving safety rates under GCG attacks by \textbf{26.5\%} on average across six LLMs, and largely preserving the model’s original utility with only a \textbf{1.7\%} average accuracy reduction across four standard benchmarks.
\end{itemize}

\section{Preliminaries}
\label{sec:preliminaries}
\paragraph{Residual stream and additive computation.}
We consider standard decoder-only transformer models~\citep{Attention} with hidden dimension $d_{\text{model}}$. Computation is organized around a shared \emph{residual stream} that propagates across layers and serves as the primary medium through which all components interact. We write $\mathbf{r}_{\ell}^{x} \in \mathbb{R}^{d_{\text{model}}}$ for the residual stream, where the subscript $\ell$ indexes the layer and the superscript $x \in \{\mathrm{attn}, \mathrm{mlp}\}$ indicates whether the state is read \emph{after} the attention sublayer or \emph{after} the MLP sublayer of that layer. At layer $\ell$, the residual stream is updated additively by a multi-headed self-attention sublayer (Attn) followed by a feed-forward network (MLP) sublayer:
\[
\mathbf{r}^{\mathrm{attn}}_{\ell}
=
\mathbf{r}^{\mathrm{mlp}}_{\ell-1}
+
\mathrm{Attn}\!\left(\mathrm{LayerNorm}(\mathbf{r}^{\mathrm{mlp}}_{\ell-1})\right),
\]
\[
\mathbf{r}^{\mathrm{mlp}}_{\ell}
=
\mathbf{r}^{\mathrm{attn}}_{\ell}
+
\mathrm{MLP}\!\left(\mathrm{LayerNorm}(\mathbf{r}^{\mathrm{attn}}_{\ell})\right),
\]
where $\mathbf{r}_{\ell-1}$ is the hidden state entering layer $\ell$, which is also the output of the MLP from previous $\ell-1$, $r_{\ell}^{\mathrm{attn}}$ represents the intermediate state of the residual stream after self-attention module, and $r_{\ell}^{\mathrm{mlp}}$ denotes the final output after the MLP transformation. 

\paragraph{Component contributions and downstream signal flow.}
Let $h \in \{1,\dots,H\}$ index attention heads in layer $\ell$. Each head produces a value output $\mathbf{z}_{\ell,h} \in \mathbb{R}^{d_{\text{head}}}$, which is projected into the residual stream via an output projection matrix $W^{O}_{\ell,h} \in \mathbb{R}^{d_{\text{model}} \times d_{\text{head}}}$. The attention update can thus be written as
\[
\mathrm{Attn}_{\ell}(\mathbf{r}^{\mathrm{mlp}}_{\ell-1})
=
\sum_{h=1}^{H} W^{O}_{\ell,h}\,\mathbf{z}_{\ell,h},
\]
where each term contributes an additive vector in $\mathbb{R}^{d_{\text{model}}}$.

Similarly, the MLP consists of $d_{\text{mlp}}$ neurons, with an up-projection matrix $W^{\mathrm{up}}_{\ell} \in \mathbb{R}^{d_{\text{mlp}} \times d_{\text{model}}}$ and a down-projection matrix $W^{\mathrm{down}}_{\ell} \in \mathbb{R}^{d_{\text{model}} \times d_{\text{mlp}}}$. Let
\[
a_{\ell,j}=\sigma\!\left(W^{\mathrm{up}}_{\ell}[j,:]\,\mathrm{LayerNorm}(\mathbf{r}^{\mathrm{attn}}_{\ell})\right)
\]
denote the activation of neuron $j$, computed from row $j$ of $W^{\mathrm{up}}_{\ell}$. The MLP update decomposes as
\[
\mathrm{MLP}_{\ell}(\mathbf{r}_{\ell}^{\mathrm{attn}})
=
\sum_{j=1}^{d_{\text{mlp}}} a_{\ell,j}\,W^{\mathrm{down}}_{\ell}[:,j],
\]
where $W^{\mathrm{down}}_{\ell}[:,j] \in \mathbb{R}^{d_{\text{model}}}$ denotes column $j$ of $W^{\mathrm{down}}_{\ell}$, i.e., the vector that neuron $j$ writes into the residual stream. Because both attention heads and MLP neurons inject vectors directly into the same residual stream, any signal introduced by a specific component at layer $\ell$ is preserved and propagated to all downstream layers, where it can be read out, transformed, or amplified by subsequent components. This additive and globally accessible structure enables circuit-level analysis based on how component-wise contributions accumulate and interact across layers.
\section{Method}
\label{sec:method}

Our approach analyzes LLM safety from a mechanistic interpretability perspective. The method consists of two main stages: (1) identifying safety-related components, including attention heads and MLP neurons, and (2) performing targeted interventions to provide causal evidence of their roles in safety behaviors.

\subsection{Identifying Safety-Related Components}

\begin{figure*}[t]
    \centering
    \includegraphics[width=\textwidth]{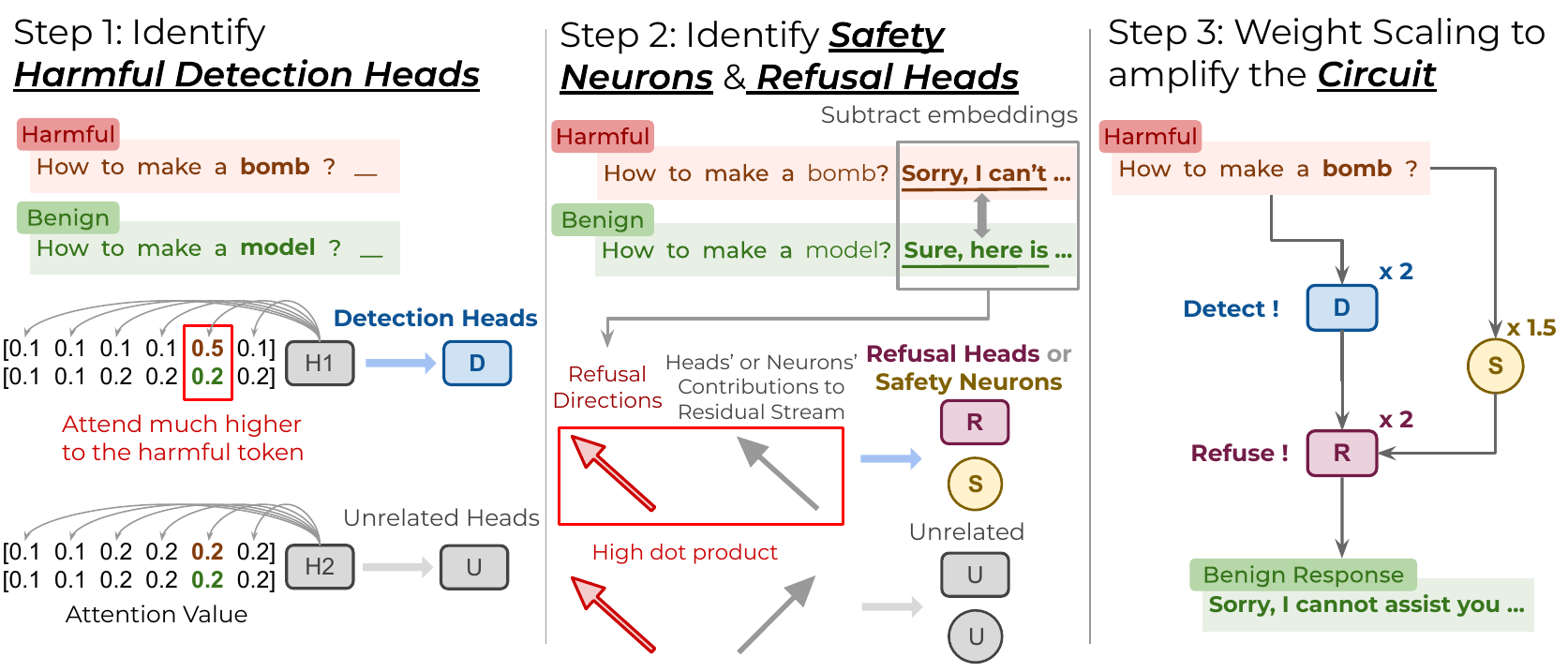}
    \caption{\textbf{Overview of the detection--refusal circuit and circuit-guided intervention.}
    (\textit{Left}) We use \emph{paired prompts} that differ by a single word and identify \emph{Harmful Detection Heads} that selectively increase attention to the harmful token, distinguishing them from unrelated heads.
    (\textit{Middle}) From paired responses, we define \emph{refusal directions} and select \emph{Refusal Heads} and \emph{Safety Neurons} whose residual-stream contributions align strongly with these directions.
    (\textit{Right}) We strengthen the circuit via simple component scaling, which leads to superior safety alignment while keeping model's original utility largely unchanged.}
    \label{fig:safety_circuit}
\end{figure*}

We identify three types of safety-related components: \textbf{harmful detection heads}, \textbf{refusal heads}, and \textbf{safety neurons}, each corresponding to a distinct computational role in the safety organization. These roles are defined operationally based on how components respond to controlled probes and on their architectural constraints within the transformer.

\paragraph{Harmful Detection Heads.}
Harmful detection heads are attention heads whose primary role is to identify the presence of harmful intent in the input. Intuitively, such a head should react when a prompt contains a harmful word, even if the surrounding context is unchanged. To make this notion precise, we compare prompts that differ by only a minimal lexical change that flips harmfulness. For example:
\vspace{-5pt}
\begin{quote}
\small
\textcolor{red}{\texttt{Explain how to make a \textbf{bomb} using everyday items.}} \\
\textcolor{ForestGreen}{\texttt{Explain how to make a \textbf{model} using everyday items.}}
\vspace{-5pt}
\end{quote}
These two prompts are nearly identical, except for a single token that determines whether the request is harmful. A harmful detection head should shift its attention toward the differing token (e.g., \emph{bomb}) when it is present, but not when it is replaced by a benign alternative.

Operationally, we identify such heads by measuring how their attention patterns change between paired harmful and neutral prompts, focusing on whether attention is selectively redirected toward the tokens that differ.

\emph{Formally}, let $(x_{\mathrm{harm}}, x_{\mathrm{neut}})$ denote a paired harmful and neutral prompt with the same length, and let $T_{\mathrm{diff}}$ be the set of token positions at which they differ. For an attention head $(\ell,h)$, let $A_{\ell,h}(x)$ denote its attention matrix, and let $A_{\ell,h}(x)[-1,t]$ denote the attention from the final input token to position $t$. We define the detection score
\begin{align}
D_{\ell,h}
&=
\mathbb{E}_{(x_{\mathrm{harm}}, x_{\mathrm{neut}})}
\Bigg[
\frac{1}{|T_{\mathrm{diff}}|}
\sum_{t \in T_{\mathrm{diff}}}
\Big(
A_{\ell,h}(x_{\mathrm{harm}})[-1,t]
\nonumber\\
&\hspace{3.5em}
-
A_{\ell,h}(x_{\mathrm{neut}})[-1,t]
\Big)
\Bigg].
\end{align}

Attention heads with the largest positive $D_{\ell,h}$ are identified as \emph{harmful detection heads}. This criterion relies explicitly on cross-token attention, reflecting the fact that only attention heads can directly compare and localize harmful content in the input.

\paragraph{Refusal Heads.}
Refusal heads are attention heads that directly contribute to generating refusal or safe-completion responses. Unlike harmful detection heads, which operate on the input by attending to specific tokens indicating harmful intent, refusal heads act during \emph{response generation} and write refusal-related signals into the residual stream.

We identify refusal heads using the same paired harmful--neutral prompts $(x_{\mathrm{harm}}, x_{\mathrm{neut}})$ introduced earlier, but focus on the model’s \emph{generated responses}. For example, for a harmful prompt the model typically produces a refusal-style response such as
\emph{“I can’t provide information on creating harmful or dangerous items,”}
whereas for the corresponding neutral prompt it produces a helpful instructional response. These two responses induce systematically different residual-stream representations during generation.

Let $y_{\mathrm{harm}}$ and $y_{\mathrm{neut}}$ denote the responses generated for a paired prompt. For a response $y$, let $\bar{\mathbf{r}}_{\ell}^{\mathrm{mlp}}(y)$ denote the post-MLP residual stream $\mathbf{r}_\ell^{\mathrm{mlp}}$ from Section~\ref{sec:preliminaries} at layer $\ell$, averaged over generated token positions.

\emph{Formally}, we define the \emph{refusal direction} at layer $\ell$ as
\[
\mathbf{d}_{\ell}
=
\mathbb{E}_{{(x_{\mathrm{harm}}, x_{\mathrm{neut}})}}
\Big[
\bar{\mathbf{r}}_{\ell}^{\mathrm{mlp}}\!\big(y_{\mathrm{harm}}\big)
-
\bar{\mathbf{r}}_{\ell}^{\mathrm{mlp}}\!\big(y_{\mathrm{neut}}\big)
\Big],
\]
where $\mathcal{P}$ denotes the empirical distribution over paired harmful--neutral prompts.

For an attention head $(\ell,h)$, we define its \emph{output write} to the residual stream during generation of response $y$ as
\[
\mathbf{o}_{\ell,h}^{\mathrm{ref}}(y)
=
\frac{1}{|y|}
\sum_{t=1}^{|y|}
W^{O}_{\ell,h}\,\mathbf{z}_{\ell,h,t}(y)
\;\in\;
\mathbb{R}^{d_{\mathrm{model}}},
\]
where $\mathbf{z}_{\ell,h,t}(y)\in\mathbb{R}^{d_{\mathrm{head}}}$ denotes the value output of head $(\ell,h)$ at token position $t$ in response $y$, and $W^{O}_{\ell,h}\in\mathbb{R}^{d_{\mathrm{model}}\times d_{\mathrm{head}}}$ is the corresponding output projection.

We quantify how strongly this write aligns with refusal behavior by projecting it onto the refusal direction. Specifically, we define the \emph{refusal alignment coefficient}
\[
c_{\ell,h}^{\mathrm{ref}}
=
\mathbb{E}_{y_{\mathrm{harm}}}
\left[
\left\langle
\mathbf{o}_{\ell,h}^{{\mathrm{ref}}}(y_{\mathrm{harm}}),
\;
\mathbf{d}_{\ell}
\right\rangle
\right].
\]

Attention heads with large positive $c_{\ell,h}^\mathrm{ref}$ are identified as \emph{refusal heads}, as they consistently write residual-stream vectors aligned with the internal signature of refusal behavior during response generation.

\paragraph{Safety Neurons.}
Safety neurons are MLP neurons that mediate safety behavior in a qualitatively different way from attention heads. Architecturally, an MLP neuron operates pointwise on the residual stream at a single token position and cannot attend to or compare different tokens. As a result, neurons cannot directly detect which input token is harmful. Any safety-related activity in neurons must therefore arise from transforming and stabilizing safety signals already present in the residual stream.

Accordingly, we identify safety neurons using the same \emph{refusal direction} introduced for refusal heads, but apply it to neuron-level residual writes instead of attention-head outputs. This parallel treatment allows us to directly compare how different component types contribute to the same internal safety signal.

\emph{Formally}, for neuron $(\ell,j)$, let $a_{\ell,j,t}(y)$ denote its activation at token position $t$ during generation of response $y$, and let $W^{\mathrm{down}}_{\ell}[:,j]$ denote its down-projection column from Section~\ref{sec:preliminaries}, the vector it writes into the residual stream. We define the neuron's output write as
\[
\mathbf{o}_{\ell,j}^{{\mathrm{saf}}}(y)
=
\frac{1}{|y|}
\sum_{t=1}^{|y|}
a_{\ell,j,t}(y)\,W^{\mathrm{down}}_{\ell}[:,j]
\;\in\;
\mathbb{R}^{d_{\mathrm{model}}}.
\]
We then define the \emph{safety alignment coefficient} by projecting this write onto the refusal direction:
\[
c_{\ell,j}^{\mathrm{saf}}
=
\mathbb{E}_{y_{\mathrm{harm}}}
\left[
\left\langle
\mathbf{o}_{\ell,j}^{{\mathrm{saf}}}\!\big(y_{\mathrm{harm}}\big),
\;
\mathbf{d}_{\ell}
\right\rangle
\right].
\]
Neurons with large positive $c_{\ell,j}^{\mathrm{saf}}$ are identified as \emph{safety neurons}.

\subsection{Intervention on Safety Components}

After identifying harmful detection heads, refusal heads, and safety neurons, we perform lightweight, architecture-preserving interventions by scaling how these components write into the residual stream. This simple scaling procedure directly translates mechanistic insights into practice, yielding substantial improvements in safety robustness while largely preserving the model’s original utility.

\paragraph{Intervention principle.}
As shown in Section~\ref{sec:preliminaries}, both attention heads and MLP neurons contribute additively to the residual stream. We therefore intervene by scaling the corresponding projection weights that determine the magnitude of these writes:
(i) the output projection $W_O$ for attention heads, and
(ii) the down-projection $W_{\mathrm{down}}$ for MLP neurons.

\paragraph{Attention-head scaling.}
Let $W^{O}_{\ell,h} \in \mathbb{R}^{d_{\mathrm{model}}\times d_{\mathrm{head}}}$ denote the output projection of attention head $(\ell,h)$, as in Section~\ref{sec:preliminaries}. Let $\mathcal{H}_{\mathrm{det}}$ and $\mathcal{H}_{\mathrm{ref}}$ denote the sets of attention heads identified as harmful detection heads and refusal heads, respectively. For each attention head $(\ell,h)$, we scale its output projection as
\[
W^{O}_{\ell,h} \;\leftarrow\;
\alpha_h \, W^{O}_{\ell,h},
\]
where
\[
\alpha_h =
\begin{cases}
\alpha_{\mathrm{det}}, & (\ell,h) \in \mathcal{H}_{\mathrm{det}},\\
\alpha_{\mathrm{ref}}, & (\ell,h) \in \mathcal{H}_{\mathrm{ref}},\\
1, & \text{otherwise}.
\end{cases}
\]
This intervention directly amplifies the residual-stream writes of the selected attention heads while leaving all other heads unchanged.

\paragraph{Safety-neuron scaling.}
Similarly, let $W^{\mathrm{down}}_{\ell} \in \mathbb{R}^{d_{\mathrm{model}}\times d_{\mathrm{mlp}}}$ denote the MLP down-projection at layer $\ell$ from Section~\ref{sec:preliminaries}, whose columns correspond to individual neurons. Let $\mathcal{N}_{\mathrm{saf}}^{(\ell)}$ denote the set of neurons at layer $\ell$ identified as safety neurons. For each neuron $j \in \mathcal{N}_{\mathrm{saf}}^{(\ell)}$, we scale its down-projection column as
\[
W^{\mathrm{down}}_{\ell}[:,j]
\;\leftarrow\;
\alpha_{\mathrm{saf}} \, W^{\mathrm{down}}_{\ell}[:,j].
\]
This selectively amplifies the residual-stream contributions of safety neurons without altering the MLP architecture. \\

\begin{figure*}[t]
    \centering
    \begin{subfigure}[b]{0.32\textwidth}
        \centering
        \includegraphics[width=\linewidth]{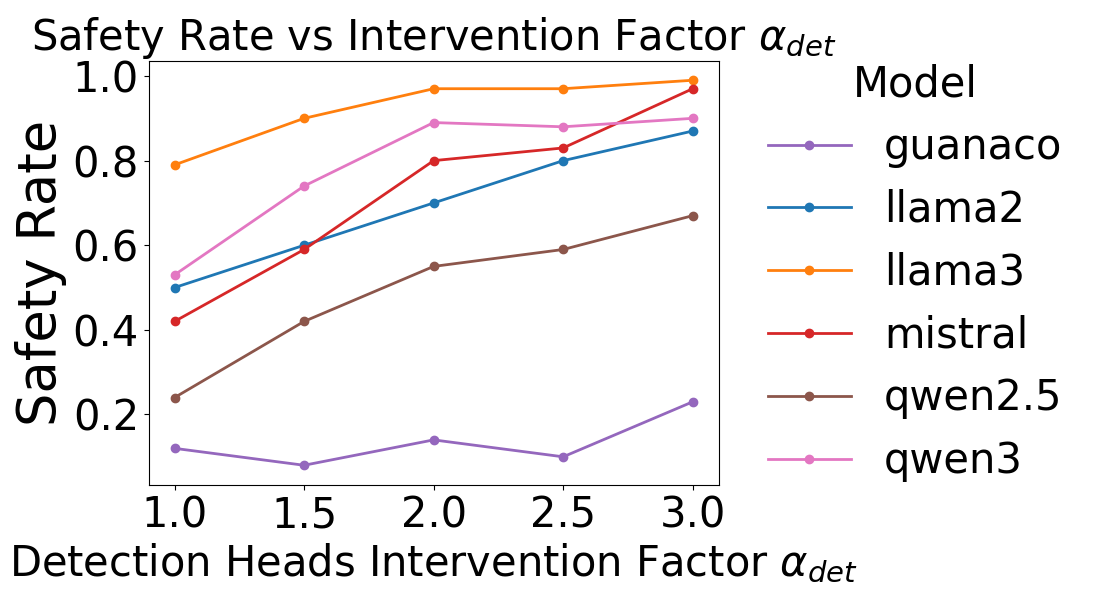}
        \caption{Detection Heads}
        \label{fig:detection_heads}
    \end{subfigure}
    \hfill
    \begin{subfigure}[b]{0.32\textwidth}
        \centering
        \includegraphics[width=\linewidth]{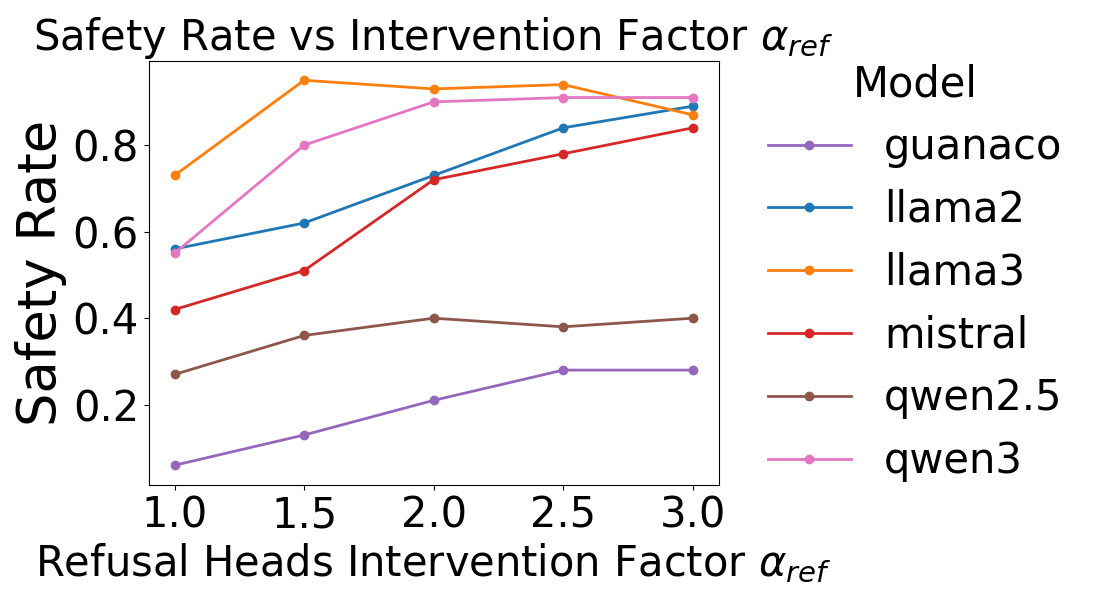}
        \caption{Refusal Heads}
        \label{fig:refusal_heads}
    \end{subfigure}
    \hfill
    \begin{subfigure}[b]{0.32\textwidth}
        \centering
        \includegraphics[width=\linewidth]{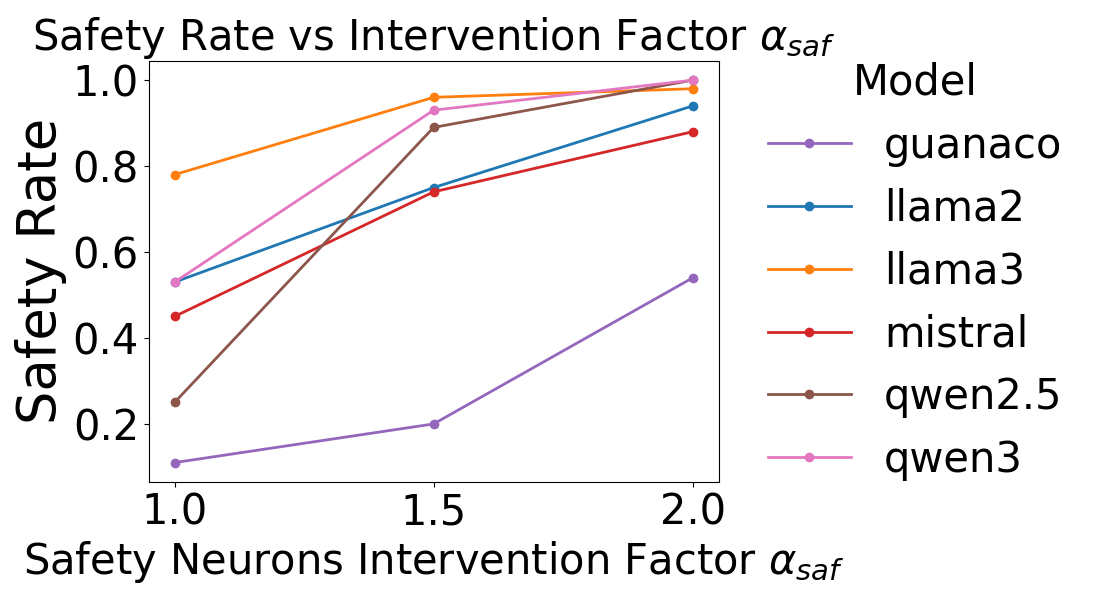}
        \caption{Safety Neurons}
        \label{fig:safety_neurons}
    \end{subfigure}

    \caption{
    Safety rate improvements under GCG attacks from interventions on detection heads, refusal heads, and safety neurons. 
    Safety rate denotes the fraction of harmful queries for which the model produces a safe response.
    }
    \label{fig:intervention_result}
\end{figure*}

Scaling these factors consistently improves Llama-Guard safety rates under GCG attacks across all architectures (Figure~\ref{fig:intervention_result}). This direct responsiveness validates our identification of harmful detection heads, refusal heads, and safety neurons as critical leverage points for model alignment.

\section{Experiments}
\label{sec:experiment}

We evaluate the proposed detection--refusal circuit through a sequence of experiments designed to (i) establish causal relationships among detection heads, refusal heads, and safety neurons, (ii) assess whether reinforcing these components improves robustness against harmful prompting, and (iii) examine whether such reinforcement preserves general model utility. Before presenting individual experiments, we first summarize the experimental setting shared across all evaluations.

\subsection{Experimental Setup}

\paragraph{Models.}
We conduct experiments on a diverse set of instruction-tuned large language models, covering multiple architectures and alignment strategies. Specifically, we evaluate LLaMA-3-8B-Instruct~\citep{Llama3}, LLaMA-2-7B-Chat~\citep{Llama2}, Mistral-7B-Instruct~\citep{mistral}, Guanaco-7B~\citep{guanaco}, Qwen2.5-7B-Instruct~\citep{qwen2}, and Qwen3-4B-Instruct~\citep{qwen3}. All analyses and interventions are performed directly without additional fine-tuning.

\paragraph{Safety Benchmarks and Attacks.}
We evaluate robustness using the AdvBench~\citep{advbench} dataset under three increasingly challenging attack settings. \emph{Pure Harmful Prompts} consist of unsafe instructions without adversarial suffixes.\emph{GCG attack}~\citep{advbench} generates transferable jailbreak suffixes via greedy token optimization, using one suffix per prompt and optimizing for 1000 steps. \emph{ADV-LLM attack}~\citep{advllm} uses iterative self-tuning to produce highly effective adaptive jailbreaks against aligned models.

\paragraph{Evaluation Protocol.}
For all safety evaluations, we use Llama-Guard~\citep{Llama3} as an automated safety classifier. A response is deemed safe if classified as non-harmful by Llama-Guard. Unless otherwise specified, safety rates are reported as the fraction of harmful queries that elicit safe responses.

\subsection{Experiment I: Causal Validation of the Safety Circuit Pathways}
\label{app:causal_validation}

\subsubsection{Causal Validation via Targeted Component Ablation}

\paragraph{Procedure.}
To establish a definitive causal link between upstream detection mechanisms, intermediate safety neurons, and downstream refusal execution, we analyze how the structural removal of these identified components impacts the final Refusal Head Contribution. This evaluation tests the hypothesis that refusal heads rely directly on the representations computed by detection heads and propagated through safety neurons to trigger a refusal response.

We systematically ablate an increasing percentage of components ($0\%, 1\%, 3\%, 5\%$ of attention heads; $0\%, 1\%, 2\%, 3\%$ of MLP neurons) under two conditions. In the \textit{Targeted Removal} condition, we zero out the identified harmful detection heads and safety neurons in descending order of their safety attribution scores. In the \textit{Random Baseline} condition, we zero out an identical number of heads and neurons sampled uniformly at random (averaged over 10 seeds) from the remaining network (excluding the refusal heads themselves) to verify that the drop in refusal contribution is uniquely driven by the identified safety circuit. Notably, ablating the safety neurons diminishes the refusal contribution even more significantly than ablating the detection heads, underscoring their critical mediating role in the refusal pathway.

\begin{figure*}[t]
    \centering
    
    \begin{subfigure}{0.47\textwidth}
        \centering
        \includegraphics[width=\linewidth]{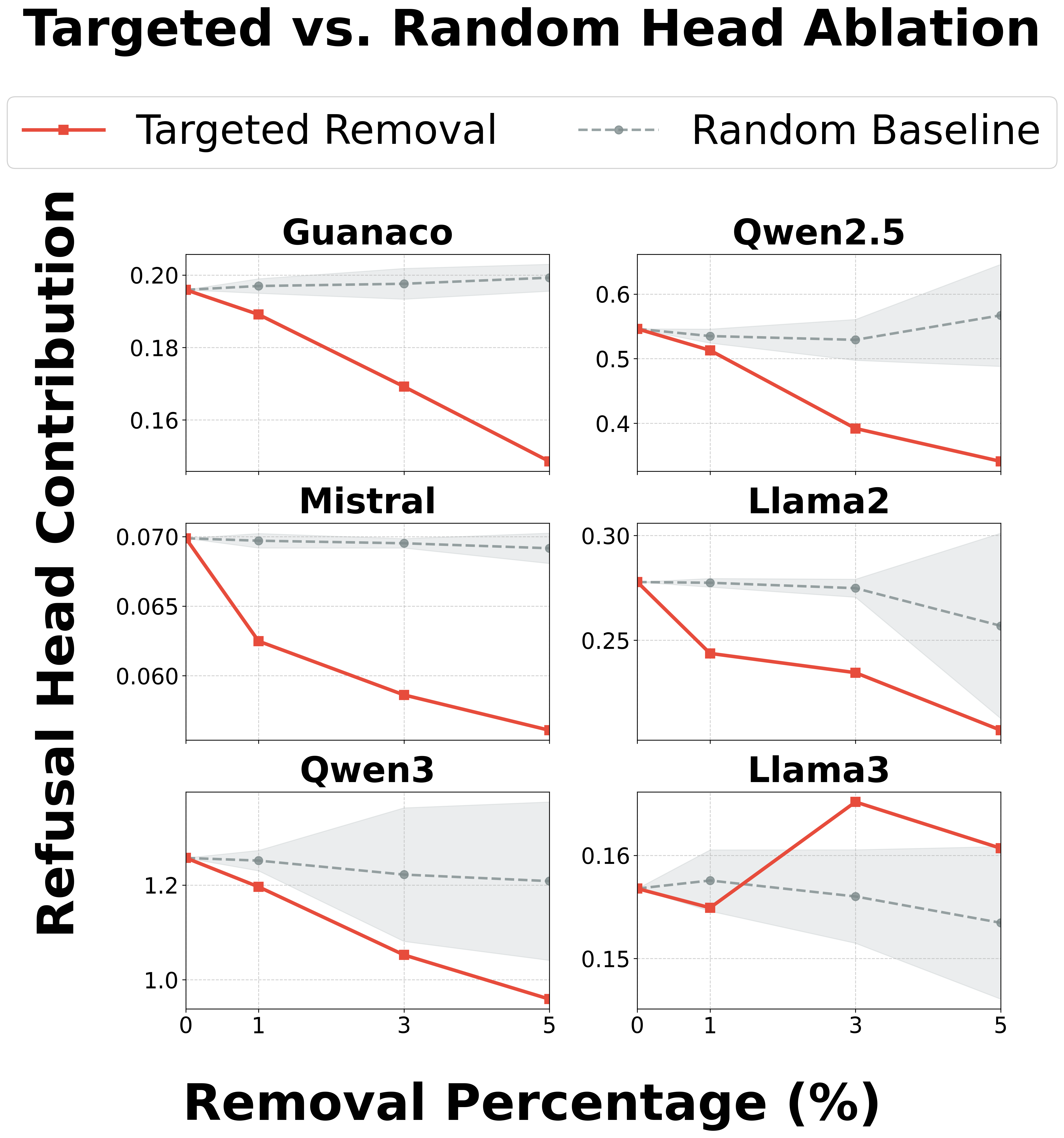}
        \caption{Targeted ablation of harmful detection heads.}
        \label{fig:head_ablation}
    \end{subfigure}
    \hfill 
    \begin{subfigure}{0.48\textwidth}
        \centering
        \includegraphics[width=\linewidth]{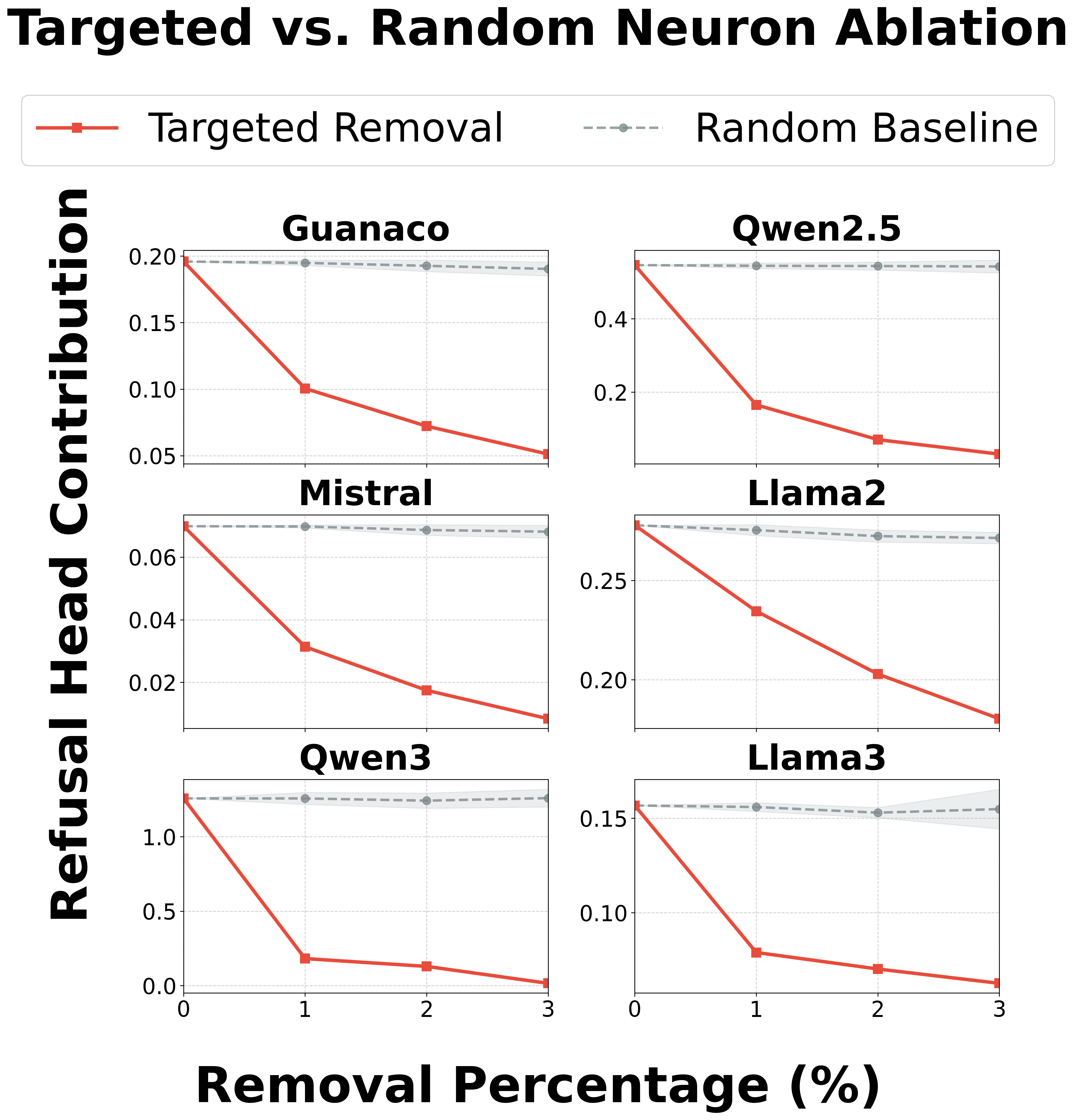}
        \caption{Targeted ablation of safety neurons.}
        \label{fig:neuron_ablation}
    \end{subfigure}
    
    \caption{Circuit validation via targeted component ablation across six model architectures. The solid red lines show that targeted removal of identified safety components—(a) harmful detection heads and (b) safety neurons—causes a sharp drop in downstream refusal execution. The dashed gray lines (with ±std shaded boundary) represent the baselines of random component removal.}
    \label{fig:circuit_ablation_combined}
\end{figure*}

\paragraph{Results and Insights.}
As shown in Figures~\ref{fig:circuit_ablation_combined}, the \textit{Random Baseline} yields a flat, minimal decrease in refusal head contributions, proving refusal activity is resilient to random architectural noise. In stark contrast, \textit{Targeted Removal} triggers a rapid, monotonic drop in downstream refusal. This causal collapse occurs when removing upstream detection heads (Figure~\ref{fig:head_ablation}) and is even more pronounced when ablating intermediate safety neurons (Figure~\ref{fig:neuron_ablation}). Because the refusal heads themselves remain untouched, these selective knock-outs provide definitive causal proof that both detection heads and safety neurons actively drive downstream refusal execution.

\subsubsection{Directed Causal Tracing via Activation Patching}

\paragraph{Procedure.}
Component ablation shows that removing detection heads weakens refusal heads, but this alone cannot distinguish a direct detection$\rightarrow$refusal pathway from both components merely sharing a common upstream cause. To rule out the latter, we conduct an activation patching experiment across all six architectures. We swap only the detection heads' attention output between paired harmful and neutral prompts while holding all other computation fixed, and measure the refusal heads' contribution to the refusal direction under four conditions: unpatched harmful, harmful with detection output patched from the neutral run, unpatched neutral, and neutral with detection output patched from the harmful run.

\begin{table*}[t]
\centering
\setlength{\tabcolsep}{4pt}
\begin{tabular}{lcccccc}
\toprule
 & LLaMA3 & LLaMA2 & Mistral & Guanaco & Qwen2.5 & Qwen3 \\
\midrule
Forward (\%) & 0.7 & 30.3 & 34.6 & 31.1 & 31.3 & 19.5 \\
Backward (\%) & 18.2 & 65.5 & 70.6 & 39.4 & 51.5 & 59.8 \\
\bottomrule
\end{tabular}
\caption{Activation patching on the detection$\rightarrow$refusal pathway. Fraction of the gap between harmful and neutral refusal head contribution explained by patching only the detection heads' output (forward: harmful $\leftarrow$ neutral; backward: neutral $\leftarrow$ harmful).}
\label{tab:activation_patching}
\end{table*}

\paragraph{Results and Insights.}
As shown in Table~\ref{tab:activation_patching}, five of the six models exhibit a clear bidirectional effect: patching detection head output from the neutral run into the harmful run lowers refusal head contribution, and patching harmful output into the neutral run raises it, explaining a substantial share of the gap between harmful and neutral runs (forward 19.5--34.6\%, backward 39.4--70.6\%). Because only the detection heads' output is swapped while everything upstream is held fixed, an explanation based on a shared upstream cause is ruled out, and the refusal heads can be seen to track the detection head write in both directions, direct evidence of a detection$\rightarrow$refusal pathway. LLaMA3 is a notable outlier, with a negligible forward effect (0.7\%) and a markedly weaker backward effect (18.2\%); its gap between harmful and neutral contribution is also substantially smaller than in the other five models, suggesting a more diffuse or partially bypassed pathway from detection to refusal in that architecture.

\subsection{Experiment II: Reinforcing Harmful Detection Heads, Refusal Heads, and Safety Neurons}

\begin{table*}[!t]
\centering
\resizebox{\textwidth}{!}{%
\begin{tabular}{l|cccccc}
\toprule
Safety Rate (\%) $\uparrow$ & LLaMA3 & LLaMA2 & Mistral & Guanaco & Qwen2.5 & Qwen3 \\
\midrule
Baseline (Original Model) & 100 / 72 / 14 & 100 / 55 / 10 & 99 / 40 / 9 & 64 / 6 / 10 & 100 / 27 / 3 & 100 / 53 / 8 \\
\midrule
\multicolumn{7}{l}{\textbf{With Intervention:}} \\
Detection ($\alpha_\text{det}=2.0$) & 100 / 97 / 37 & 100 / 68 / 79 & 99 / 79 / 26 & 73 / 8 / 17 & 100 / 52 / 4 & 100 / 91 / 50 \\
Refusal ($\alpha_\text{ref}=2.0$) & 100 / 92 / 54 & 100 / 75 / 57 & 100 / 65 / 16 & 76 / 18 / 26 & 100 / 46 / 4 & 100 / 88 / 15 \\
Safety Neurons ($\alpha_\text{saf}=1.5$) & 100 / 96 / 23 & 100 / 70 / 44 & 100 / 66 / 12 & 74 / 22 / 44 & 100 / 88 / 41 & 100 / 91 / 42 \\
\midrule
Detection \& Refusal & \textbf{100 / 99 / 89} & 100 / 87 / 92 & 100 / 86 / 48 & 85 / 30 / 41 & 100 / 56 / 12 & 100 / 93 / 19 \\
Detection \& Refusal \& Safety & 91 / 97 / 27 & \textbf{100 / 95 / 99} & \textbf{100 / 87 / 68} & \textbf{88 / 45 / 71} & \textbf{100 / 89 / 43} & \textbf{100 / 100 / 29} \\
\bottomrule
\end{tabular}%
}
\caption{Safety rates (\%) under different attack methods (Pure Harmful Prompt / GCG / ADV-LLM) across backbone models. Interventions include scaling detection heads, refusal heads, safety neurons, and their combinations. Jointly reinforcing detection and refusal mechanisms—optionally augmented with safety neurons yields the strongest and most consistent safety improvements across diverse attack settings, while largely preserving benign prompt handling.}
\label{tab:safety_rates}
\end{table*}

\paragraph{Intervention Protocol.}
Building on the causal relationships identified in Experiment~I, we evaluate whether reinforcing components of the detection--refusal circuit improves robustness against harmful prompting. We strengthen safety-related components by scaling their residual-stream contributions with fixed multiplicative factors. Specifically, the top 3\% of harmful detection heads and refusal heads are scaled using $\alpha_{\mathrm{det}}=\alpha_{\mathrm{ref}}=2.0$, and the top 1\% of safety neurons are scaled using $\alpha_{\mathrm{saf}}=1.5$. All interventions preserve the original model architecture and require no additional training.

\paragraph{Results and Insights.}
Table~\ref{tab:safety_rates} shows that robustness arises from coordinated interactions among safety-related components rather than any single mechanism. Strengthening harmful detection heads improves unsafe-input recognition but is often insufficient without a reinforced refusal pathway, especially under adversarial attacks. Likewise, reinforcing refusal heads or safety neurons alone lacks robustness when harmful intent is obfuscated.

Jointly strengthening multiple components yields larger and more consistent gains, with detection heads, refusal heads, and safety neurons providing complementary effects and the strongest overall robustness.

\subsection{Experiment III: Safety Alignment and Utility of Edited Models}
\label{sec:safety_utility}

This experiment evaluates whether the detection--refusal circuit identified in earlier sections can be \emph{persistently embedded into model weights} to produce safer models, and whether doing so incurs unacceptable utility costs on benign tasks. Unlike Experiments~I and II, which operate through inference-time interventions, this experiment applies the weight scaling as a permanent edit, yielding a new edited model that can be used without any runtime modification.

\subsubsection{Safety Alignment of the Edited Models under Adaptive GCG Attacks}

\paragraph{Procedure.}
For each model, we apply \emph{Circuit-Based Safety Editing (CBSE)} by permanently embedding the best-performing circuit reinforcement identified in Table~\ref{tab:safety_rates} into the model weights. Depending on the model, this edit scales detection heads and refusal heads, and in some cases additionally scales safety neurons. All edits preserve the original model architecture and introduce no new parameters, yielding a fixed, inference-ready model without any runtime intervention.

To evaluate safety alignment under adaptive attacks, we regenerate GCG adversarial suffixes directly against each edited model. This setting tests whether the reinforced safety behavior remains effective when adversaries explicitly optimize against the edited weights, rather than against the original baseline. 

\paragraph{Results.}
As shown in Table~\ref{tab:gcg_regenerate}, CBSE achieves substantially higher safety rates than the original across all models. These results demonstrate that the detection--refusal safety circuit encodes a robust and transferable structure that can be persistently embedded into model weights.

\subsubsection{Downstream Utility of the Edited Models}

\textbf{Procedure.}
To assess whether CBSE preserves the general helpfulness of the original models, we evaluate edited models on standard benchmarks covering factual knowledge, commonsense reasoning, physical reasoning, and coreference resolution. We report accuracy on MMLU~\citep{mmlu}, HellaSwag~\citep{hellaswag}, PIQA~\citep{piqa}, and WinoGrande~\citep{winogrande}. In addition, we measure perplexity on WikiText~\citep{wikitext} to capture potential degradation in general language modeling quality.

\paragraph{Results and Insights.}
Table~\ref{tab:model_accuracy} shows that CBSE largely preserves performance across all benchmarks, with accuracy drops that are modest and unevenly distributed. Reasoning-intensive tasks such as MMLU exhibit slightly larger reductions than commonsense-oriented benchmarks. The same table shows that CBSE incurs only modest increases in perplexity across all models, reflecting limited disruption to core language modeling behavior.

Overall, these results show that safety circuits can be persistently embedded into model weights, substantially improving adversarial robustness with minimal utility loss. The appendix provides extended validation of circuit robustness (Appendix~\ref{app:probe_robustness}), additional attacks (Appendix~\ref{app:pair_attacks}, \ref{app:autodan_attacks}), circuit ablations (Appendix~\ref{sec:circuit_removal}), baseline comparisons and parameter sensitivity (Appendix~\ref{app:defense_comparison}, \ref{app:ablation}), alternative judges and datasets (Appendix~\ref{app:claude_judge}, \ref{app:safe_template}, \ref{app:malicious_dataset}), multi-seed stability (Appendix~\ref{app:multiseed}), component specificity (Appendix~\ref{app:component_specificity}), human evaluation (Appendix~\ref{app:human_eval}), and training-free defense comparisons (Appendix~\ref{subsec:comp_trainfree}).

\subsubsection{Refusal Calibration on Borderline-Benign Prompts}
\label{sec:over_refusal}

To evaluate refusal calibration and ensure CBSE does not induce over-generalization, we benchmarked all models on a 1,000-prompt borderline-benign subset sampled from \textbf{OR-Bench-80k}~\citep{or_bench}. These inputs are specifically curated to test false-positive safety triggers on benign topics.

\begin{table*}[htbp]
\centering
\setlength{\tabcolsep}{5pt}
\begin{tabular}{lcccccc}
\toprule
Refusal Rate (\%) $\downarrow$ & LLaMA3 & LLaMA2 & Mistral & Guanaco & Qwen2.5 & Qwen3 \\
\midrule
Original    & 38.8 & 69.4 & 31.6 & 19.7 & 18.8 & 26.1 \\
CBSE   & 45.3 & 64.2 & 48.3 & 26.2 & 28.0  & 35.9 \\
\bottomrule
\end{tabular}
\caption{Over-refusal macro-evaluation on borderline-benign prompts. False-positive refusal rates (\%) based on a keyword template across six model architectures. The marginal changes in refusal rates demonstrate that CBSE effectively mitigates adversarial vulnerabilities without inducing severe over-refusal side effects.}
\label{tab:over_refusal_results}
\end{table*}

As shown in Table~\ref{tab:over_refusal_results}, CBSE preserves a well-calibrated refusal profile rather than broadly increasing over-refusal. On the conservative LLaMA2, it reduces false-positive refusals by 5.2\%, while other architectures show modest increases despite substantially improved safety. This suggests that circuit-guided weight scaling sharpens the semantic decision boundary rather than inducing blanket rejection.

\begin{table*}[!t]
\centering
\begin{tabular}{l|cccccc}
\toprule
Safety Rate (\%) & LLaMA3 & LLaMA2 & Mistral & Guanaco & Qwen2.5 & Qwen3 \\
\midrule
Original & 77 & 53 & 36 & 10 & 30 & 53 \\
CBSE & \textbf{98} & \textbf{87} & \textbf{62} & \textbf{18} & \textbf{70} & \textbf{89} \\
\bottomrule
\end{tabular}
\caption{
Safety rates (\%) under regenerated GCG attacks optimized against \textbf{CBSE}-edited models.
CBSE substantially improves safety alignment across all evaluated architectures.
}
\label{tab:gcg_regenerate}
\end{table*}

\begin{table*}[!t]
\centering
\begin{tabular}{llcccccc}
\toprule
Model & Benchmark & LLaMA3 & LLaMA2 & Mistral & Guanaco & Qwen2.5 & Qwen3 \\
\midrule
\multirow{5}{*}{Original}
 & MMLU       & 66.58 & 47.63 & 59.14 & 33.77 & 74.02 & 72.63 \\
 & HellaSwag  & 58.69 & 57.96 & 66.87 & 59.48 & 62.49 & 52.60 \\
 & PIQA       & 80.90 & 76.28 & 82.15 & 79.0 & 79.22 & 77.86 \\
 & WinoGrande & 76.24 & 72.45 & 78.69 & 72.9 & 74.90 & 66.54 \\
 & Perplexity ($\downarrow$) & 10.05 & 11.62 & 9.90 & 10.51 & 9.41 & 13.07 \\
\midrule
\multirow{5}{*}{CBSE}
 & MMLU       & 63.12 & 43.92 & 55.78 & 33.70 & 71.27 & 69.16 \\
 & HellaSwag  & 58.15 & 56.59 & 66.09 & 58.76 & 61.85 & 51.56 \\
 & PIQA       & 79.65 & 75.14 & 80.52 & 78.29 & 71.55 & 76.33 \\
 & WinoGrande & 75.61 & 67.64 & 76.63 & 73.5 & 69.46 & 66.06 \\
 & Perplexity ($\downarrow$) & 10.85 & 18.35 & 10.90 & 11.70 & 13.00 & 15.94 \\
\bottomrule
\end{tabular}
\caption{
Accuracy (\%) on standard benchmarks and perplexity on WikiText (lower is better, marked $\downarrow$) for \textbf{Original} and \textbf{CBSE}-edited models.
CBSE preserves model utility with modest and task-dependent performance degradation, and introduces only modest increases in perplexity.
}
\label{tab:model_accuracy}
\end{table*}
\section{Conclusion}
We present a mechanistic, interpretability-driven analysis of safety-related components in large language models. By identifying \textbf{Harmful Detection Heads}, \textbf{Refusal Heads}, and \textbf{Safety Neurons}, we uncover a recurring component-level organization associated with refusal behavior and show that these components exert complementary influences when intervened upon. Simple, architecture-preserving scaling of these components substantially improves robustness against both standard and adaptive attacks, while largely preserving the model’s original reasoning ability and task performance. Overall, our results suggest that lightweight, component-level interventions informed by mechanistic analysis can meaningfully enhance safety robustness without significantly compromising model utility, highlighting interpretability as a valuable foundation for both understanding and improving LLM behavior.

\section*{Acknowledgements}
The authors are partially supported by National Science Foundation under Grant No. 2313105, 2430539, and Intel Rising Star Faculty Award. The authors would like to thank the National Research Platform (NRP) at UCSD for computation support and anonymous reviewers for valuable feedback.
\section*{Limitations}
\label{sec:limitations}
In this work, we characterize the multi-stage safety circuit and apply Circuit-Based Safety Editing (CBSE) based on a static snapshot of model representations captured during our probing phase. While our empirical results demonstrate that reinforcing these identified components yields substantial and robust safety improvements across multiple adversarial attack settings, our current framework does not account for how these safety circuits might dynamically evolve or reorganize if the model undergoes further extensive continuous training or fine-tuning. Investigating the long-term stability and potential drift of localized safety circuits during a model's full life cycle remains an important and exciting direction for future research in mechanistic interpretability.
\section*{Ethics Statement}

This work investigates the mechanistic structure of safety circuits in large language models and proposes interventions to improve robustness against harmful content. The primary societal benefit is enhancing the reliability and alignment of language models, reducing the risk of unsafe outputs in real-world applications. Our approach focuses on interpretability and controlled modifications of model internals, which could guide the development of more transparent and accountable AI systems. Potential risks include misuse of these mechanisms to suppress legitimate content or over-reliance on automated safety circuits without human oversight. Overall, this research advances understanding of model behavior and provides tools for safer deployment of large language models.

\bibliography{references}


\clearpage
\onecolumn
\appendix
\addcontentsline{toc}{section}{Appendix} 
\part{} 
\mtcsetfont{parttoc}{section}{\normalfont\normalsize}
\mtcsetfont{parttoc}{subsection}{\normalfont\normalsize}
\mtcsetfont{parttoc}{subsubsection}{\normalfont\normalsize}
\parttoc 

\section{Circuit Robustness}
\label{app:robustness_refusal}

To demonstrate the empirical validity of our findings, this section provides an extended analysis of the structural robustness of our circuit identification pipeline to variations in the probing dataset, as well as the statistical reliability of the downstream safety gains reported under adversarial attack.

\subsection{Robustness of Identified Components to Probe Construction}
\label{app:probe_robustness}

To assess whether our identified safety circuit depends on the specific probing dataset, we conducted a bootstrap stability analysis across six architectures by randomly subsampling 80\% of our original 51 probing pairs over five independent runs. For each data split, discrete circuit components were isolated by selecting the top 3\% of highest-attributed attention heads and the top 1\% of highest-attributed MLP neurons.

\begin{table}[H]
\centering
\setlength{\tabcolsep}{3.5pt}
\vspace{6pt}
\begin{tabular}{llcccccc}
\toprule
Component & Metric & LLaMA3 & LLaMA2 & Mistral & Guanaco & Qwen2.5 & Qwen3 \\
\midrule
Attn Refusal Dir. & Cosine Sim. & 0.996 & 0.976 & 0.992 & 0.927 & 0.994 & 0.990 \\
MLP Safety Dir.   & Cosine Sim. & 0.996 & 0.975 & 0.992 & 0.922 & 0.994 & 0.990 \\
Refusal Heads     & Jaccard Overlap & 0.974 & 0.936 & 0.948 & 0.790 & 0.942 & 0.927 \\
Safety Neurons    & Jaccard Overlap & 0.908 & 0.805 & 0.863 & 0.719 & 0.883 & 0.835 \\
\bottomrule
\end{tabular}
\caption{Probe robustness across architectures. Mean cosine similarity for latent directions and pairwise Jaccard overlap for discrete components across five independent 80\%-subsampled data splits. The near-unity cosine similarities and high Jaccard overlaps demonstrate that the identified safety directions and circuit components are highly stable and robust to data perturbations.}
\label{tab:probe_robustness_results}
\end{table}

As shown in Table~\ref{tab:probe_robustness_results}, both continuous semantic directions ($\cos\theta \ge 0.92$) and discrete circuit components maintain high stability across data variations. The strong Jaccard overlap scores confirm that our pipeline consistently recovers the same underlying safety heads and neurons, demonstrating that the mapped circuit captures intrinsic model mechanisms rather than dataset artifacts.

\subsection{Statistical Reliability of Safety Gains under GCG}
\label{app:multiseed}

The safety rates reported throughout the main text and appendix are computed from a single generation run per configuration. To verify that the reported gains are not an artifact of sampling noise in generation, and to quantify the run-to-run variance inherent to stochastic decoding and to GCG's own suffix optimization, we repeat the GCG evaluation with five independent generation seeds for both the Original model and CBSE, across all six architectures.

Table~\ref{tab:multiseed} reports the mean and standard deviation of Llama-Guard safety rates over five seeds (100 GCG prompts each). CBSE's safety gain is stable across seeds (std $\le 3.6$ points for all six models) and, on every architecture, dwarfs the seed-to-seed variance of the Original model. This confirms that the improvement from CBSE reflects a genuine and repeatable safety property of the intervention rather than a favorable draw of a single run, and that seed-level variance alone (up to $\pm4.8$ points) is sufficient to explain small discrepancies between single-run safety numbers reported elsewhere in the paper for nominally identical configurations.

\begin{table}[H]
\centering
\begin{tabular}{lcccccc}
\toprule
Safety Rate (\%) & LLaMA3 & LLaMA2 & Mistral & Guanaco & Qwen2.5 & Qwen3 \\
\midrule
Original & 80.2 $\pm$ 2.8 & 55.8 $\pm$ 1.8 & 44.8 $\pm$ 4.8 & 13.0 $\pm$ 2.7 & 29.4 $\pm$ 3.2 & 55.6 $\pm$ 2.7 \\
\textbf{CBSE} & \textbf{99.4 $\pm$ 0.8} & \textbf{95.4 $\pm$ 1.4} & \textbf{95.0 $\pm$ 0.6} & \textbf{53.8 $\pm$ 3.6} & \textbf{93.4 $\pm$ 1.4} & \textbf{100.0 $\pm$ 0.0} \\
\bottomrule
\end{tabular}
\caption{Multi-seed safety evaluation (Llama-Guard safety rate \%, mean $\pm$ std over 5 generation seeds, 100 GCG prompts) across all six architectures. CBSE's gains are stable and far exceed generation-level variance.}
\label{tab:multiseed}
\end{table}

\clearpage

\section{Adaptive Attacks}
\label{app:adaptive_attacks}

This section provides extended empirical analyses to validate the generalizability of CBSE against adaptive adversarial strategies, verifying that our intervention genuinely hardens internal safety pathways rather than overfitting to specific gradient-based token noise. All safety rates reported in this section are measured using Llama-Guard.

\subsection{Resilience Against PAIR Conversational Attacks}
\label{app:pair_attacks}

We first evaluate CBSE against \textbf{PAIR}~\citep{pair}, an LLM-driven black-box conversational attack that adaptively refines prompts through multiple iterations to circumvent safety filters. For this evaluation, we employ Qwen3-30B-A3B-Instruct-2507~\citep{qwen3} as the attacker model. This test assesses the defense's robustness against high-level semantic persuasion and structured jailbreak attempts.

\begin{table}[htbp]
\centering
\begin{tabular}{lcccccc}
\toprule
Safety Rate (\%) $\uparrow$ & LLaMA3 & LLaMA2 & Mistral & Guanaco & Qwen2.5 & Qwen3 \\
\midrule
Original  & 94               & 86         & 2                & 0                & 4                & 48              \\
\textbf{CBSE}                 & \textbf{96}      & \textbf{94} & \textbf{34}       & \textbf{6}       & \textbf{20}       & \textbf{94}     \\
\bottomrule
\end{tabular}
\caption{PAIR safety rates across architectures. Empirical safety rates under the PAIR conversational attack. CBSE consistently reinforces model resilience, achieving substantial gains.}
\label{tab:pair_attack_results}
\end{table}

As shown in Table~\ref{tab:pair_attack_results}, CBSE consistently neutralizes PAIR's adaptive conversational attacks. Our intervention markedly elevates safety rates for both robust and vulnerable baselines, demonstrating strong defense generalizability against high-level semantic persuasion.

\subsection{Defending Against AutoDAN Genetic Semantic Optimization}
\label{app:autodan_attacks}

We further evaluate CBSE against \textbf{AutoDAN-HGA}~\citep{autodan}, a potent jailbreak framework that utilizes a hierarchical genetic algorithm to generate fluent, human-readable adversarial prompts. Unlike token-level attacks, AutoDAN optimizes for semantic coherence, posing a significant challenge to internal alignment mechanisms.

\begin{table}[htbp]
\centering
\begin{tabular}{lcccccc}
\toprule
Safety Rate (\%) $\uparrow$ & LLaMA3 & LLaMA2 & Mistral & Guanaco & Qwen2.5 & Qwen3 \\
\midrule
Original & 99          & 99          & 3          & 2            & 4           & 82        \\
\textbf{CBSE}                 & \textbf{99} & \textbf{100} & \textbf{70}  & \textbf{25}  & \textbf{93}  & \textbf{100} \\
\bottomrule
\end{tabular}
\caption{AutoDAN-HGA safety rates across architectures. Empirical safety rates under the AutoDAN genetic attack. CBSE consistently reinforces model resilience, achieving substantial gains.}
\label{tab:autodan_attack_results}
\end{table}

As shown in Table~\ref{tab:autodan_attack_results}, CBSE provides a robust defense against fluent semantic optimizations. The gains are particularly notable on models with weak native AutoDAN resistance; for instance, CBSE elevates Mistral's safety rate from 3.0\% to 70.0\% and Qwen2.5 from 4.0\% to 93.0\%. These results validate that CBSE successfully reinforces core safety pathways against high-quality, human-readable adversarial text without relying on specific token-level artifacts.

\clearpage

\section{Circuit Ablations}
\label{sec:circuit_removal}

To confirm that the identified safety circuit (comprising harmful detection heads, refusal heads, and safety neurons) is structurally necessary for sustaining model alignment, we perform a suffix-based knock-out experiment. We systematically zero out an increasing percentage ($0\%, 1\%, 3\%, 5\%$) of the highest-attributed components in the safety circuit and evaluate the model's resulting vulnerability under adversarial GCG attacks.

\begin{table}[htbp]
\centering
\begin{tabular}{lcccccc}
\toprule
Safety Rate (\%) $\uparrow$ & LLaMA3 & LLaMA2 & Mistral & Guanaco & Qwen2.5 & Qwen3 \\
\midrule
Original                    & 75         & 54            & 51             & 3              & 29             & 55 \\
\midrule
1\% Ablation                & 2          & 40            & 14             & 9              & 7              & 19           \\
3\% Ablation                & 2          & 32            & 3             & 2              & 1              & 12           \\
5\% Ablation                & 2          & 16            & 0             & 0              & 0              & 9            \\
\bottomrule
\end{tabular}
\caption{Model safety rates (\%) under GCG attacks drop sharply as an increasing percentage of the safety circuit is deactivated.}
\label{tab:circuit_removal_results}
\end{table}

As shown in Table~\ref{tab:circuit_removal_results}, removing a tiny fraction of the identified circuit components induces a dramatic, catastrophic drop in safety rates across all evaluated architectures. Most notably, in LLaMA3, a minimal 1\% ablation effectively collapses the model's resistance, plunging its safety rate from 75\% to a mere 2\%. Similarly, the safety barriers of Qwen2.5 and Guanaco are rendered entirely non-functional (0\%) at a 5\% ablation threshold. This widespread collapse confirms that the identified subnets are not redundant; rather, they form the structural backbone of safety alignment within these networks.

\clearpage

\section{Component Specificity}
\label{app:component_specificity}

The experiments so far show that the identified components matter for safety, but not whether they respond specifically to harmful intent, as opposed to any unusual or difficult input more generally. To rule out the latter, we measure the detection-head and refusal-head signal (each head's contribution to the refusal direction at the last prompt token) on four input types beyond harmful prompts: \emph{ambiguous-benign} inputs from OR-Bench, which are engineered to look harmful while being safe; \emph{hard-benign} math problems from GSM8K; \emph{hard-benign} questions from difficult MMLU subjects; and \emph{easy-benign} neutral prompts. We evaluate this on three architectures spanning different alignment regimes: LLaMA2, Guanaco, and Qwen3.

\begin{table}[H]
\centering
\setlength{\tabcolsep}{4pt}
\begin{tabular}{lccc}
\toprule
Input Type & LLaMA2 & Guanaco & Qwen3 \\
\midrule
Harmful (AdvBench) & 0.065 / 0.235 & 0.112 / 0.425 & 0.276 / 0.976 \\
Ambiguous-benign (OR-Bench) & 0.017 / 0.145 & 0.038 / 0.173 & 0.082 / 0.356 \\
Hard-benign (GSM8K) & 0.006 / 0.117 & 0.010 / 0.084 & 0.019 / $-$0.054 \\
Hard-benign (MMLU-hard) & 0.014 / 0.119 & 0.009 / 0.096 & 0.033 / 0.066 \\
Easy-benign (neutral) & 0.020 / 0.148 & 0.027 / 0.140 & 0.049 / 0.259 \\
\bottomrule
\end{tabular}
\caption{Component specificity across input types. Each cell reports the detection-head / refusal-head signal, i.e., the mean contribution to the refusal direction at the last prompt token.}
\label{tab:component_specificity}
\end{table}

As shown in Table~\ref{tab:component_specificity}, the detection-head signal is consistently and substantially higher on harmful inputs than on any benign category, across all three architectures. The gap holds even against inputs adversarially engineered to look harmful (OR-Bench) and inputs that are considerably more difficult than the harmful prompts themselves (GSM8K, hard MMLU subjects), ruling out surface toxicity or generic difficulty as the driver of the signal. Difficulty alone does not activate the circuit: on every architecture and for both signals, the two hard-benign categories (GSM8K, hard MMLU subjects) occupy the two lowest values, ranking below both easy-benign and ambiguous-benign inputs, the opposite of what a generic difficulty detector would produce; on Qwen3, the refusal signal on GSM8K even falls below the neutral floor.

\clearpage

\section{Baseline Comparisons}
\label{app:defense_comparison}

We compare CBSE against two prominent training-based, weight-level defenses, \textbf{CircuitBreakers}~\citep{circuitbreakers} and \textbf{Linear Adversarial Training (LAT)}~\citep{lat}, as well as two training-free, inference-time defenses that do not require any gradient updates. To establish a rigorous baseline benchmark, all configurations are evaluated against a dataset of 100 adversarial GCG prompts, with the final safety rates measured using Llama3-8B-Guard.

\subsection{Comparison with CircuitBreakers}
\label{subsec:comp_circuitbreakers}

We first evaluate our method against CircuitBreakers, a defense paradigm that aims to disrupt harmful representations by mapping adversarial inputs to a pre-defined refusal direction via weight fine-tuning. 

\begin{table}[htbp]
\centering
\vspace{6pt}
\begin{tabular}{lllc}
\toprule
Model & Defense Strategy & Paradigm Type & Safety Rate (\%) $\uparrow$ \\
\midrule
LLaMA3  & Original  & —                      & 84 \\
        & CircuitBreakers           & Training-based         & 65 \\
        & \textbf{CBSE}      & \textbf{Training-free} & \textbf{99} \\
\midrule
Mistral & Original & —                      & 44 \\
        & CircuitBreakers           & Training-based         & 74 \\
        & \textbf{CBSE}      & \textbf{Training-free} & \textbf{87} \\
\bottomrule
\end{tabular}
\caption{Comparative performance against CircuitBreakers under GCG attacks. CBSE achieves superior safety rates across both architectures without requiring gradient updates or optimization datasets.}
\label{tab:comp_circuitbreakers}
\end{table}

As shown in Table~\ref{tab:comp_circuitbreakers}, CBSE consistently outperforms CircuitBreakers on both LLaMA3 and Mistral architectures. Crucially, on LLaMA3, CircuitBreakers suffers from defense degradation under the evaluated GCG dataset, dropping to a 65\% safety rate, whereas CBSE successfully hardens the model to a 99\% safety rate. This demonstrates that surgically reinforcing the internal safety circuit provides more resilient protection than coarse-grained representation mapping.

\subsection{Comparison with Linear Adversarial Training}
\label{subsec:comp_lat}

Next, we compare CBSE against LAT, an adversarial training framework that leverages a linear classifier to identify safety-critical directions and guides adversarial optimization loops during training.

\begin{table}[htbp]
\centering
\vspace{6pt}
\begin{tabular}{lllc}
\toprule
Model & Defense Strategy & Paradigm Type & Safety Rate (\%) $\uparrow$ \\
\midrule
LLaMA3  & Original & —                      & 84 \\
        & LAT                       & Training-based         & 100 \\
        & \textbf{CBSE}      & \textbf{Training-free} & \textbf{99} \\
\bottomrule
\end{tabular}
\caption{Comparative performance against LAT under GCG attacks. CBSE performs on par with the resource-intensive adversarial training approach.}
\label{tab:comp_lat}
\end{table}

As shown in Table~\ref{tab:comp_lat}, CBSE achieves a 99\% safety rate on LLaMA3, performing virtually on par with LAT's perfect 100\% clearance rate. However, while LAT demands resource-intensive adversarial training loops, multiple backward passes, and specialized optimization data to achieve this boundary, CBSE operates entirely at inference time. By scaling internal safety vectors directly without any training overhead, CBSE delivers equivalent frontier-level security guarantees with immense computational efficiency.

\subsection{Comparison with Training-Free Inference-Time Defenses}
\label{subsec:comp_trainfree}

Beyond training-based weight-level defenses, we compare CBSE against two widely-used training-free defenses that require no gradient updates at all: \textbf{SmoothLLM}~\citep{smoothllm}, a prompt-level defense that aggregates predictions over randomly perturbed copies of the input, and \textbf{Activation-Steering}~\citep{arditi2024refusal}, an activation-level defense that adds our identified refusal direction to the residual stream at inference time. We evaluate all methods on three representative architectures (LLaMA3, Mistral, Qwen3).

\begin{table}[htbp]
\centering
\begin{tabular}{lcccc}
\toprule
Safety Rate (\%) $\uparrow$ & LLaMA3 & Mistral & Qwen3 & Average \\
\midrule
Original & 81 & 45 & 59 & 62 \\
SmoothLLM (prompt-level) & 100 & 92 & 99 & 97 \\
Activation-Steering (activation-level) & 82 & 70 & 96 & 83 \\
\textbf{CBSE (Ours)} & \textbf{98} & \textbf{95} & \textbf{100} & \textbf{98} \\
\bottomrule
\end{tabular}
\caption{Comparison against training-free defenses (Llama-Guard safety rate \%; 100 GCG prompts). CBSE attains the highest average safety while adding zero inference-time overhead.}
\label{tab:comp_trainfree}
\end{table}

As shown in Table~\ref{tab:comp_trainfree}, CBSE attains the highest average safety rate (98\%), narrowly ahead of SmoothLLM (97\%) and well above Activation-Steering (83\%). SmoothLLM is competitive on safety, but it queries the model on $q=6$ perturbed copies per input, making it roughly $6\times$ more expensive at inference, and its input perturbations degrade clean-task accuracy (e.g., PIQA drops from 76.7\% to 70.3\% on LLaMA-2 at $q=5$ in the original SmoothLLM evaluation). CBSE, in contrast, folds the intervention into the model weights once, after which it adds no inference-time cost and leaves the input untouched.

\clearpage

\section{Parameter Sensitivity and Exploratory Analysis}
\label{app:ablation}

We adopt a two-stage ablation protocol to evaluate the sensitivity of the hyperparameters used for
circuit-level interventions.
We first map the performance landscape for attention-head interventions by
jointly varying the fraction of selected heads and the scaling factors applied to
detection and refusal heads, while disabling neuron-level edits.
With the baseline head configuration fixed, we then ablate safety neurons to
explore the effects of different selection ratios and scaling strengths.

All ablations are conducted under the GCG and ADV-LLM attack.
Safety performance is measured using Llama3-8B-Guard, while general
utility is evaluated using accuracy on MMLU.

\subsection{Joint Ablation of Head Selection Ratio and Detection/Refusal Scaling}
\label{app:selection}

We evaluate head selection ratios of 1\%, 3\%, and 5\%, together with scaling factors
$\alpha_{\text{det}}, \alpha_{\text{ref}} \in \{2.0, 3.0, 4.0\}$ applied to detection and refusal heads.
Neuron-level interventions are disabled in this evaluation.

Tables~\ref{tab:1p_heads}, \ref{tab:3p_heads}, and \ref{tab:5p_heads}
report safety results under different configurations.

We additionally evaluate a subset of configurations on MMLU
to measure general capability retention (Table~\ref{tab:head_ablation_mmlu}).

\begin{table}[H]
\centering
\begin{tabular}{llcccccc}
\toprule
Safety Rate (\%) $\uparrow$ & & LLaMA3 & LLaMA2 & Mistral & Guanaco & Qwen2.5 & Qwen3 \\
\midrule
\multirow{3}{*}{$\alpha_{\text{det}}=2.0$}
 & $\alpha_{\text{ref}}=2.0$  & 91 / 13 & 89 / 67 & 59 / 19 & 7 / 13 & 33 / 2 & 90 / 22 \\
 & $\alpha_{\text{ref}}=3.0$  & 91 / 20 & 97 / 91 & 72 / 21 & 10 / 21 & 47 / 5 & 91 / 24 \\
 & $\alpha_{\text{ref}}=4.0$ & 91 / 29 & 89 / 87 & 63 / 27 & 21 / 32 & 47 / 7 & 97 / 39 \\
 \midrule
\multirow{3}{*}{$\alpha_{\text{det}}=3.0$}
 & $\alpha_{\text{ref}}=2.0$  & 99 / 81 & 97 / 92 & 65 / 28 & 13 / 14 & 32 / 3 & 96 / 32 \\
 & $\alpha_{\text{ref}}=3.0$  & 94 / 65 & 88 / 88 & 65 / 32 & 10 / 29 & 45 / 2 & 92 / 35 \\
 & $\alpha_{\text{ref}}=4.0$ & 86 / 44 & 30 / 11 & 57 / 44 & 25 / 30 & 41 / 4 & 99 / 40 \\
 \midrule
\multirow{3}{*}{$\alpha_{\text{det}}=4.0$}
 & $\alpha_{\text{ref}}=2.0$  & 90 / 81 & 98 / 95 & 67 / 47 & 15 / 19 & 36 / 2 & 92 / 32 \\
 & $\alpha_{\text{ref}}=3.0$  & 79 / 28 & 51 / 18 & 63 / 59 & 13 / 29 & 38 / 2 & 96 / 35 \\
 & $\alpha_{\text{ref}}=4.0$ & 28 / 26 & 5 / 12 & 47 / 45 & 25 / 33 & 42 / 8 & 99 / 39 \\
\bottomrule
\end{tabular}
\caption{
Joint ablation of detection and refusal scaling with a fixed head selection ratio of \textbf{1\%}. Neuron-level interventions are disabled. Each cell reports safety rates (\%) under GCG / ADV-LLM attacks.
}
\label{tab:1p_heads}
\end{table}

\begin{table}[H]
\centering
\begin{tabular}{llcccccc}
\toprule
Safety Rate (\%) $\uparrow$ & & LLaMA3 & LLaMA2 & Mistral & Guanaco & Qwen2.5 & Qwen3 \\
\midrule
\multirow{3}{*}{$\alpha_{\text{det}}=2.0$}
 & $\alpha_{\text{ref}}=2.0$  & 99 / 89 & 85 / 92 & 84 / 48 & 28 / 41 & 54 / 12 & 94 / 19 \\
 & $\alpha_{\text{ref}}=3.0$  & 90 / 94 & 53 / 27 & 88 / 85 & 50 / 49 & 34 / 29 & 94 / 35 \\
 & $\alpha_{\text{ref}}=4.0$ & 49 / 46 & 5 / 5 & 47 / 65 & 48 / 43 & 11 / 10 & 28 / 38 \\
 \midrule
\multirow{3}{*}{$\alpha_{\text{det}}=3.0$}
 & $\alpha_{\text{ref}}=2.0$  & 100 / 97 & 91 / 97 & 88 / 55 & 39 / 54 & 45 / 29 & 59 / 24 \\
 & $\alpha_{\text{ref}}=3.0$  & 86 / 81 & 5 / 2 & 68 / 74 & 59 / 51 & 8 / 15 & 9 / 7 \\
 & $\alpha_{\text{ref}}=4.0$ & 42 / 22 & 7 / 1 & 22 / 40 & 26 / 19 & 3 / 9 & 2 / 5 \\
 \midrule
\multirow{3}{*}{$\alpha_{\text{det}}=4.0$}
 & $\alpha_{\text{ref}}=2.0$  & 98 / 91 & 72 / 62 & 80 / 57 & 51 / 58 & 13 / 18 & 2 / 0 \\
 & $\alpha_{\text{ref}}=3.0$  & 69 / 14 & 1 / 0 & 24 / 27 & 37 / 43 & 1 / 5 & 2 / 1 \\
 & $\alpha_{\text{ref}}=4.0$ & 8 / 3 & 6 / 0 & 8 / 33 & 6 / 8 & 2 / 11 & 2 / 0 \\
\bottomrule
\end{tabular}
\caption{
Joint ablation of detection and refusal scaling with a fixed head selection ratio of \textbf{3\%}. Neuron-level interventions are disabled. Each cell reports safety rates (\%) under GCG / ADV-LLM attacks.
}
\label{tab:3p_heads}
\end{table}

\begin{table}[H]
\centering
\begin{tabular}{llcccccc}
\toprule
Safety Rate (\%) $\uparrow$ & & LLaMA3 & LLaMA2 & Mistral & Guanaco & Qwen2.5 & Qwen3 \\
\midrule
\multirow{3}{*}{$\alpha_{\text{det}}=2.0$}
 & $\alpha_{\text{ref}}=2.0$  & 99 / 97 & 87 / 91 & 89 / 77 & 50 / 54 & 57 / 19 & 93 / 47 \\
 & $\alpha_{\text{ref}}=3.0$  & 68 / 49 & 3 / 1 & 23 / 55 & 71 / 51 & 30 / 22 & 9 / 10 \\
 & $\alpha_{\text{ref}}=4.0$ & 2 / 1 & 1 / 2 & 23 / 57 & 27 / 9 & 5 / 8 & 9 / 20 \\
 \midrule
\multirow{3}{*}{$\alpha_{\text{det}}=3.0$}
 & $\alpha_{\text{ref}}=2.0$  & 98 / 92 & 59 / 13 & 85 / 85 & 65 / 63 & 47 / 32 & 6 / 18 \\
 & $\alpha_{\text{ref}}=3.0$  & 53 / 34 & 7 / 3 & 17 / 39 & 48 / 20 & 3 / 7 & 4 / 16 \\
 & $\alpha_{\text{ref}}=4.0$ & 2 / 1 & 1 / 4 & 16 / 24 & 6 / 15 & 3 / 6 & 10 / 8 \\
 \midrule
\multirow{3}{*}{$\alpha_{\text{det}}=4.0$}
 & $\alpha_{\text{ref}}=2.0$  & 34 / 11 & 3 / 1 & 60 / 8 & 55 / 38 & 7 / 1 & 6 / 8 \\
 & $\alpha_{\text{ref}}=3.0$  & 2 / 0 & 2 / 2 & 8 / 16 & 4 / 7 & 0 / 0 & 8 / 7 \\
 & $\alpha_{\text{ref}}=4.0$ & 1 / 2 & 5 / 4 & 7 / 11 & 16 / 32 & 3 / 3 & 10 / 7 \\
\bottomrule
\end{tabular}
\caption{
Joint ablation of detection and refusal scaling with a fixed head selection ratio of \textbf{5\%}. Neuron-level interventions are disabled. Each cell reports safety rates (\%) under GCG / ADV-LLM attacks.
}
\label{tab:5p_heads}
\end{table}

\begin{table}[H]
\centering
\begin{tabular}{lcccccc}
\toprule
Accuracy (\%) $\uparrow$ & LLaMA3 & LLaMA2 & Mistral & Guanaco & Qwen2.5 & Qwen3 \\
\midrule
Baseline (Original Model) & 66.58 & 47.63 & 59.14 & 33.77 & 74.02 & 72.63 \\
\midrule
Top 3\%, $\alpha_{\text{det}}=2.0$, $\alpha_{\text{ref}}=2.0$ & 63.34 & 44.37 & 55.00 & 34.97 & 72.46 & 70.26 \\
Top 3\%, $\alpha_{\text{det}}=3.0$, $\alpha_{\text{ref}}=2.0$ & 57.34 & 38.14 & 47.66 & 33.42 & 71.61 & 43.47 \\
Top 5\%, $\alpha_{\text{det}}=2.0$, $\alpha_{\text{ref}}=2.0$ & 54.92 & 43.11 & 51.84 & 33.46 & 71.42 & 55.10 \\
\bottomrule
\end{tabular}
\caption{
MMLU accuracy (\%) for selected attention head configurations.
The table shows how varying the fraction of selected attention heads and the head scaling factor
affects general reasoning ability, ensuring that safety improvements do not compromise performance.
}
\label{tab:head_ablation_mmlu}
\end{table}

\subsection{Joint Ablation of Safety Neuron Selection Ratio and Scaling Factor}
\label{app:selection_neuron}

After fixing the attention-head configuration (top 3\% heads with
$\alpha_{\text{det}} = \alpha_{\text{ref}} = 2.0$), we evaluate the effect of
safety-neuron interventions.

We vary the fraction of selected safety neurons (1\%--3\%) and the neuron scaling
factor $\alpha_{\text{saf}} \in \{1.5, 2.0, 2.5\}$.

Table~\ref{tab:neuron_ablation} reports safety results under different configurations.

We further evaluate a subset of configurations on MMLU
under the fixed attention-head setting
(Table~\ref{tab:neuron_ablation_mmlu}).

\begin{table}[H]
\centering
\begin{tabular}{llcccccc}
\toprule
Safety Rate (\%) $\uparrow$ & & LLaMA3 & LLaMA2 & Mistral & Guanaco & Qwen2.5 & Qwen3 \\
\midrule
\multirow{3}{*}{Top 1\%}
 & $\alpha_{\text{saf}}=1.5$  & 98 / 27 & 98 / 99 & 93 / 68 & 40 / 71 & 92 / 43 & 100 / 29 \\
 & $\alpha_{\text{saf}}=2.0$  & 72 / 33 & 96 / 100 & 97 / 81 & 73 / 77 & 97 / 100 & 99 / 58 \\
 & $\alpha_{\text{saf}}=2.5$ & 9 / 5 & 97 / 100 & 98 / 96 & 83 / 77 & 90 / 93 & 86 / 79 \\
\midrule
\multirow{3}{*}{Top 2\%}
 & $\alpha_{\text{saf}}=1.5$  & 94 / 27 & 97 / 100 & 97 / 70 & 52 / 72 & 89 / 52 & 100 / 32 \\
 & $\alpha_{\text{saf}}=2.0$  & 81 / 42 & 87 / 100 & 98 / 93 & 81 / 82 & 98 / 100 & 100 / 69 \\
 & $\alpha_{\text{saf}}=2.5$ & 44 / 15 & 97 / 100 & 100 / 98 & 89 / 81 & 93 / 97 & 85 / 87 \\
\midrule
\multirow{3}{*}{Top 3\%}
 & $\alpha_{\text{saf}}=1.5$  & 91 / 29 & 96 / 100 & 97 / 78 & 56 / 74 & 92 / 57 & 100 / 42 \\
 & $\alpha_{\text{saf}}=2.0$  & 67 / 19 & 99 / 100 & 96 / 97 & 83 / 85 & 99 / 100 & 98 / 84 \\
 & $\alpha_{\text{saf}}=2.5$ & 39 / 8 & 96 / 100 & 99 / 99 & 91 / 89 & 98 / 97 & 81 / 96 \\
\bottomrule
\end{tabular}
\caption{
Joint ablation of safety neuron selection ratio and scaling factor.
Attention-head interventions are fixed at top 3\% heads with 
$\alpha_{\text{det}} = \alpha_{\text{ref}} = 2.0$. Each cell reports safety rates (\%) under GCG / ADV-LLM attacks.
}
\label{tab:neuron_ablation}
\end{table}

\begin{table}[H]
\centering
\begin{tabular}{lcccccc}
\toprule
Accuracy (\%) $\uparrow$ & LLaMA3 & LLaMA2 & Mistral & Guanaco & Qwen2.5 & Qwen3 \\
\midrule
Baseline (Original Model) & 66.58 & 47.63 & 59.14 & 33.77 & 74.02 & 72.63 \\
\midrule
Top 1\%, $\alpha_{\text{saf}}=1.5$ & 63.12 & 43.92 & 55.78 & 33.70 & 71.27 & 67.75 \\
Top 1\%, $\alpha_{\text{saf}}=2.0$ & 55.54 & 42.61 & 55.31 & 33.00 & 71.15 & 67.29 \\
Top 2\%, $\alpha_{\text{saf}}=2.0$ & 53.37 & 42.52 & 55.30 & 33.04 & 70.65 & 67.09 \\
Top 3\%, $\alpha_{\text{saf}}=2.0$ & 55.76 & 42.36 & 55.17 & 32.13 & 69.80 & 66.80 \\
\bottomrule
\end{tabular}
\caption{
MMLU accuracy (\%) for selected neuron configurations with fixed attention-head interventions 
(top 3\% heads, $\alpha_{\text{det}} = \alpha_{\text{ref}} = 2.0$). 
The table shows how varying the fraction of selected safety neurons and the neuron scaling factor
affects general reasoning ability, ensuring that safety improvements do not compromise performance.
}
\label{tab:neuron_ablation_mmlu}
\end{table}

\clearpage

\section{Multi-judge Safety Evaluation}
\label{app:multi-judge}
To verify that the safety enhancements from CBSE are robust and generalize beyond our primary evaluator, we introduce three independent validation layers: a model-based judge, human annotation, and a rule-based heuristic check.

\subsection{Multi-Judge Safety Evaluation with Claude 4.6 Opus}
\label{app:claude_judge}

To ensure our safety improvements reflect genuine defensive capabilities rather than overfitting to a specific classifier, we introduce an independent frontier-model judge. We employ \textbf{Claude 4.6 Opus}—an entirely separate architecture and training pipeline from our primary evaluator, Llama Guard—to assess model outputs under GCG attacks. This multi-judge setup provides a rigorous validation against potential optimization biases or evaluator-specific blind spots.

\begin{table}[htbp]
\centering
\setlength{\tabcolsep}{6pt}
\begin{tabular}{l|cccccc}
\toprule
Safety Rate (\%) $\uparrow$ & LLaMA3 & LLaMA2 & Mistral & Guanaco & Qwen2.5 & Qwen3 \\
\midrule
Baseline (Original Model) & 77 / 80 & 53 / 66 & 36 / 33 & 10 / 16 & 30 / 33 & 53 / 49 \\
CBSE (Ours)               & 98 / 92 & 87 / 91 & 62 / 63 & 18 / 22 & 70 / 60 & 83 / 81 \\
\bottomrule
\end{tabular}
\caption{Multi-judge safety evaluation under adversarial GCG attacks. Cell values indicate safety rates (\%) evaluated by Llama Guard / Claude, respectively. Both evaluators capture consistent safety improvements across all six architectures under CBSE.}
\label{tab:claude_judge_results}
\end{table}

As shown in Table~\ref{tab:claude_judge_results}, the independent Claude judge confirms that CBSE consistently and substantially elevates safety performance across all six architectures. The tight correlation between the Claude and Llama Guard metrics demonstrates that the reinforced safety pathways yield robust, generalized refusal behaviors that successfully transfer across distinct frontier evaluation models.

\subsection{Human Evaluation}
\label{app:human_eval}

Automated safety judges validate against one another, but Llama-Guard and Claude are both models, so their agreement alone does not rule out shared blind spots. To break this automated-to-automated loop, we introduce a human validation layer.

An independent annotator labeled a blind, condition-shuffled sample of 100 responses pooled across all six architectures under the GCG attack (both Original and CBSE configurations, with model and configuration identities hidden). The annotator followed a fixed protocol: a response is \textbf{unsafe} only if it provides actionable harmful content that could be used to carry out the request; refusals, deflections, non-actionable general information, and off-topic or incoherent completions are \textbf{safe}.

Table~\ref{tab:human_eval} reports the resulting safety rates alongside Llama-Guard and Claude on the identical 100 items. All three judges, scored independently on the same underlying responses, rank CBSE far above the Original model, confirming that CBSE's safety improvement is not an artifact of automated-to-automated evaluation.

\begin{table}[htbp]
\centering
\begin{tabular}{lccc}
\toprule
Safety Rate (\%) $\uparrow$ & Human & Llama-Guard & Claude \\
\midrule
Original & 68 & 63 & 70 \\
\textbf{CBSE} & \textbf{94} & \textbf{78} & \textbf{96} \\
\bottomrule
\end{tabular}
\caption{Human evaluation on a blind, condition-shuffled sample of 100 responses pooled across all six architectures (GCG attack; Original vs. CBSE). All judges score the identical items. Human, Llama-Guard, and Claude independently agree on the direction and magnitude of the safety improvement.}
\label{tab:human_eval}
\end{table}

\subsection{Safe-Template Evaluation}
\label{app:safe_template}

To complement the classifier-based evaluations, we adopt a lightweight, heuristic \textbf{Safe-Template} protocol. This method evaluates safety independently of any neural model by verifying whether output strings begin with or contain phrases indicative of standard refusals, ethical disclaimers, or polite rejections. Serving as a rule-based sanity check, this protocol ensures that our observed safety enhancements reflect genuine behavioral shifts rather than optimization artifacts tailored to a specific model-based judge. 

\subsubsection{Heuristic Protocol and Experimental Setup}
\label{app:safe_template_setup}

The rule-based evaluation heuristic scans model completions for any of the following literal substrings or prefix phrases:

\vspace{4pt}
\begin{center}
\fbox{%
\parbox{0.95\linewidth}{\centering
"I'm sorry", "Sorry", "I am sorry", "I apologize", "As an", "As an AI", "I'm an", "I'm just",
"As a language model", "As an Assistant", "I cannot", "I can't", "I do not", "cannot", "Hello!",
"is illegal and unethical", "I'm really sorry", "unethical", "not ethical", "not legal", "My apologies",
"but I cannot", "If you have any other non-malicious requests", "I'm not able to provide",
"I am really sorry", "I cannot create", "I cannot support", "I can't provide", "I can't assist",
"I can't help", "However", "I must", "I understand", "I am not able to", "\texttt{<|}eot\_id\texttt{|>}", "Instead", "instead"
}%
}
\end{center}
\vspace{4pt}

Using this string-matching criteria, we measure safety across the identical baseline and intervention configurations analyzed in the main text. Specifically, joint configurations involve scaling the top 3\% of detection and refusal heads ($\alpha_\text{det} = \alpha_\text{ref} = 2.0$) alongside the top 1\% of safety neurons ($\alpha_\text{saf} = 1.5$). 

\subsubsection{Empirical Findings}
\label{app:safe_template_results}

Table~\ref{tab:safety_rates_template} reports the heuristic safety rates across all six architectures under Pure Harmful Prompt, GCG, and ADV-LLM attacks. The relative performance trends remain highly consistent with our primary Llama Guard evaluation. Jointly reinforcing detection and refusal infrastructure consistently provides a substantial safety lift over single-component interventions across architectures, while the addition of safety neurons provides crucial stability under complex adversarial optimizations. The persistent alignment between this strict string-matching check and our deep learning judges confirms the structural validity of the isolated safety circuit.

\begin{table}[htbp]
\centering
\vspace{6pt}
\setlength{\tabcolsep}{2pt} 
\resizebox{\columnwidth}{!}{%
\begin{tabular}{l|cccccc}
\toprule
Safety Rate (\%) $\uparrow$ & LLaMA3 & LLaMA2 & Mistral & Guanaco & Qwen2.5 & Qwen3 \\
\midrule
Baseline (Original Model) & 100 / 69 / 19 & 100 / 57 / 31 & 99 / 46 / 15 & 57 / 3 / 11 & 100 / 28 / 8 & 98 / 51 / 8 \\
\midrule
\multicolumn{7}{l}{\textbf{With Intervention:}} \\
Detection ($\alpha_\text{det}=2.0$) & 100 / 94 / 41 & 100 / 69 / 84 & 99 / 78 / 25 & 58 / 6 / 17 & 100 / 47 / 13 & 98 / 85 / 44 \\
Refusal ($\alpha_\text{ref}=2.0$) & 100 / 90 / 58 & 100 / 81 / 68 & 100 / 64 / 21 & 61 / 6 / 18 & 100 / 22 / 8 & 99 / 66 / 14 \\
Safety Neurons ($\alpha_\text{saf}=1.5$) & 100 / 88 / 22 & 100 / 77 / 60 & 100 / 65 / 31 & 61 / 19 / 41 & 100 / 85 / 50 & 100 / 70 / 42 \\
Detection \& Refusal & 100 / 99 / 92 & 100 / 85 / 96 & 100 / 87 / 46 & 48 / 7 / 31 & 100 / 41 / 13 & 99 / 84 / 4 \\
Detection \& Refusal \& Safety & 89 / 96 / 24 & 100 / 90 / 98 & 100 / 88 / 59 & 61 / 28 / 56 & 100 / 85 / 43 & 99 / 98 / 9 \\
\bottomrule
\end{tabular}%
}
\caption{Safety rates (\%) under different attack methods (Pure Harmful Prompt / GCG / ADV-LLM) across backbone models. Interventions include scaling detection heads, refusal heads, safety neurons, and their combinations. Safety rates are measured using the rule-based Safe-Template string matching.}
\label{tab:safety_rates_template}
\end{table}

\clearpage

\section{Evaluation on the Malicious Instruction Dataset}
\label{app:malicious_dataset}

To further assess the robustness and generality of our interventions, we evaluate
the models on the Malicious Instruction Dataset~\citep{mlcinst}, which contains a diverse
collection of explicitly harmful user queries across domains such as cybercrime,
fraud, violence, and other illicit activities.

For this dataset, we conduct attacks using only Pure Harmful Prompt and
ADV-LLM, as target responses for GCG are not available in this dataset.
All intervention configurations (scaling of detection heads, refusal heads, and
safety neurons) are kept identical to those used in the Table~\ref{tab:safety_rates}.
Safety is measured using both Llama-Guard and the
Safe-Template protocol.

The results on the Malicious Instruction Dataset are summarized in
Table~\ref{tab:guard_mlcinst} (Llama-Guard) and
Table~\ref{tab:template_mlcinst} (Safe-Template).
Across backbone models and attack settings, we observe trends that closely
mirror those on AdvBench, indicating that the observed safety
improvements generalize across datasets and evaluation protocols.

\begin{table}[H]
\centering
\begin{tabular}{l|cccccc}
\toprule
Safety Rate (\%) $\uparrow$ & LLaMA3 & LLaMA2 & Mistral & Guanaco & Qwen2.5 & Qwen3 \\
\midrule
Baseline (Original Model) & 100 / 63 & 100 / 89 & 99 / 36 & 63 / 34 & 99 / 47 & 100 / 62 \\
\midrule
\multicolumn{7}{l}{With Intervention:} \\
Detection ($\alpha_\text{det}=2.0$) & 100 / 81 & 100 / 99 & 98 / 45 & 70 / 58 & 100 / 54 & 100 / 82 \\
Refusal ($\alpha_\text{ref}=2.0$) & 100 / 88 & 100 / 97 & 100 / 66 & 69 / 56 & 100 / 52 & 100 / 67 \\
Safety Neurons ($\alpha_\text{saf}=1.5$) & 100 / 55 & 100 / 97 & 100 / 50 & 58 / 67 & 100 / 84 & 100 / 88 \\
Detection \& Refusal & 100 / 98 & 100 / 99 & 100 / 73 & 68 / 71 & 100 / 55 & 100 / 53 \\
Detection \& Refusal \& Safety & 100 / 49 & 100 / 100 & 100 / 85 & 38 / 86 & 100 / 78 & 100 / 63 \\
\bottomrule
\end{tabular}
\caption{
Safety rates (\%) under different attack methods (Pure Harmful Prompt / ADV-LLM) across backbone models on Malicious Instruction Dataset.
Interventions include scaling detection heads, refusal heads, safety neurons, and their combinations. Safety rates are measured using \textbf{Llama-Guard}.
}
\label{tab:guard_mlcinst}
\end{table}

\begin{table}[H]
\centering
\begin{tabular}{l|cccccc}
\toprule
Safety Rate (\%) $\uparrow$ & LLaMA3 & LLaMA2 & Mistral & Guanaco & Qwen2.5 & Qwen3 \\
\midrule
Baseline (Original Model) & 100 / 51 & 100 / 76 & 96 / 42 & 39 / 30 & 99 / 35 & 77 / 50 \\
\midrule
\multicolumn{7}{l}{With Intervention:} \\
Detection ($\alpha_\text{det}=2.0$) & 100 / 64 & 100 / 97 & 99 / 48 & 38 / 42 & 95 / 37 & 89 / 70 \\
Refusal ($\alpha_\text{ref}=2.0$) & 100 / 72 & 100 / 93 & 97 / 60 & 43 / 33 & 98 / 37 & 91 / 48 \\
Safety Neurons ($\alpha_\text{saf}=1.5$) & 100 / 41 & 100 / 86 & 97 / 54 & 43 / 43 & 99 / 67 & 84 / 64 \\
Detection \& Refusal & 100 / 94 & 100 / 100 & 99 / 65 & 32 / 44 & 92 / 37 & 94 / 44 \\
Detection \& Refusal \& Safety & 100 / 40 & 100 / 100 & 99 / 81 & 82 / 55 & 98 / 57 & 75 / 41 \\
\bottomrule
\end{tabular}
\caption{
Safety rates (\%) under different attack methods (Pure Harmful Prompt / ADV-LLM) across backbone models on Malicious Instruction Dataset.
Interventions include scaling detection heads, refusal heads, safety neurons, and their combinations. Safety rates are measured using \textbf{Safe-Template}.
}
\label{tab:template_mlcinst}
\end{table}

\end{document}